\documentclass[conference]{IEEEtran}
\IEEEoverridecommandlockouts
\usepackage{cite}
\usepackage{amsmath,amssymb,amsfonts}
\usepackage{algorithmic}
\usepackage{graphicx}
\usepackage{textcomp}
\usepackage{xcolor}
\def\BibTeX{{\rm B\kern-.05em{\sc i\kern-.025em b}\kern-.08em
    T\kern-.1667em\lower.7ex\hbox{E}\kern-.125emX}}
    
\usepackage{url}
\usepackage{relsize}
    
\begin{document}

\title{Interpreting and Predicting Tactile Signals for the SynTouch BioTac\\
}

\author{Yashraj S. Narang$^{1}$, Balakumar Sundaralingam$^{1}$, Karl Van Wyk$^{1}$, Arsalan Mousavian$^{1}$, Dieter Fox$^{1, 2}$ % <-this % stops a space
\thanks{$^{1}$NVIDIA Corporation, Seattle, USA.}%
\thanks{$^{2}$Paul G. Allen School of Computer Science \& Engineering, University of Washington, Seattle, USA.}%
}

\maketitle

\begin{abstract}
In the human hand, high-density contact information provided by afferent neurons is essential for many human grasping and manipulation capabilities. In contrast, robotic tactile sensors, including the state-of-the-art SynTouch BioTac, are typically used to provide low-density contact information, such as contact location, center of pressure, and net force. Although useful, these data do not convey or leverage the rich information content that some tactile sensors naturally measure. This research extends robotic tactile sensing beyond reduced-order models through 1) the automated creation of a precise experimental tactile dataset for the BioTac over a diverse range of physical interactions, 2) a 3D finite element (FE) model of the BioTac, which complements the experimental dataset with high-density, distributed contact data, 3) neural-network-based mappings from raw BioTac signals to not only low-dimensional experimental data, but also high-density FE deformation fields, and 4) mappings from the FE deformation fields to the raw signals themselves. The high-density data streams can provide a far greater quantity of interpretable information for grasping and manipulation algorithms than previously accessible.
\end{abstract}

% \begin{IEEEkeywords}
% tactile sensing, finite element analysis, deep learning
% \end{IEEEkeywords}

\section{Introduction} \label{sec:introduction}

There are three major sensing modalities used in robotic grasping and manipulation: proprioception, vision, and tactile sensing. Among these modalities, tactile sensing can provide the most direct information about the physical properties of the object during interaction, including mass, moment of inertia, stiffness, friction, surface texture, temperature, and thermal conductivity, as well as information about contact locations and forces \cite{Li2020TRO}. Such information has been widely leveraged to perform low-level perceptual and control tasks, such as object classification \cite{Yuan2018ICRA}, 2D pose estimation \cite{Bauza2020CoRL}, dynamics parameter estimation \cite{Sundaralingam2020ArXiv}, slip detection \cite{Veiga2018Haptics}, contour following \cite{Lepora2019RAL}, and peg-in-hole insertion \cite{VanWyk2018TRO}.

More broadly, tactile sensing is essential for grasping and manipulation in the presence of visual occlusions, including robot self-occlusions, deep object concavities, and environmental clutter. Canonical examples of tasks under occlusion include extracting a coin from a wallet, pulling keys out of a pocket, or rummaging through a bag \cite{Mason2018AnnRev, Billard2019Science}. Furthermore, tactile sensing is invaluable for the safe handling of brittle and delicate objects, such as eggs, biscuits, glassware, fresh fruit, human tissue during surgery, and living organisms. 

\begin{figure}[thpb]
  \centering
  \includegraphics[scale=0.27]{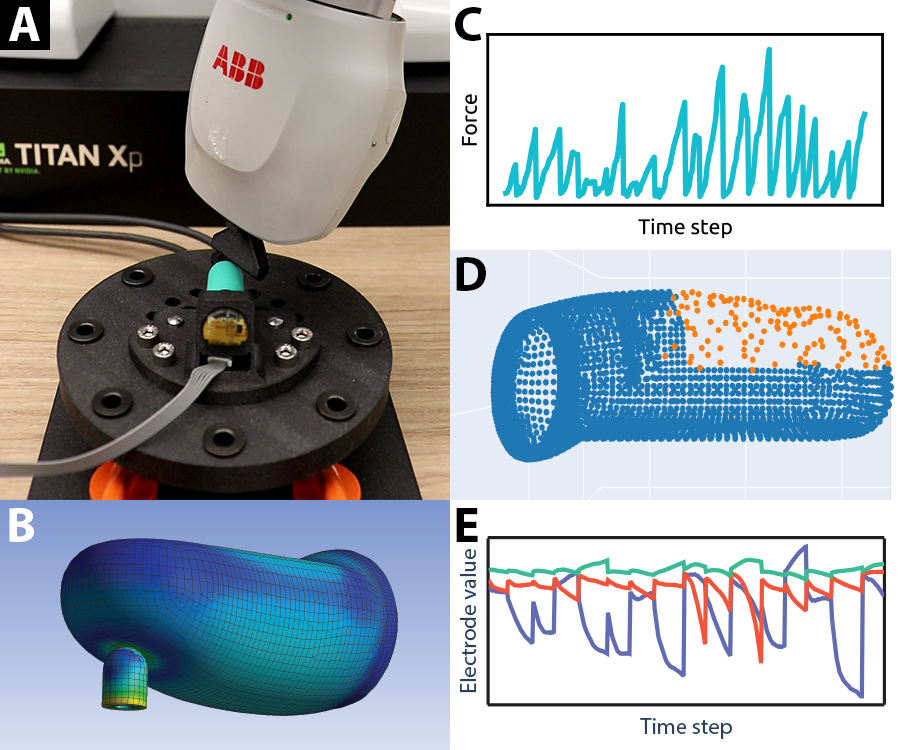}
  \caption{Paper overview. This research presents the following for the SynTouch BioTac: A) a large experimental dataset, B) the first 3D finite element model, C) regression to low-density tactile features, including 3D net force vectors, D) the first regression to high-density tactile fields, namely surface deformations of the BioTac, and E) synthesis of raw electrode signals.}
  \label{fig:intro}
\end{figure}

Fortunately, a large number of compelling tactile sensors have been developed for robotic applications, including the GelSight \cite{Yuan2017Sensors}, TacTip \cite{Ward2018SoRo}, Soft-bubble \cite{Alspach2019RoboSoft}, DIGIT \cite{Lambeta2020ICRA}, overlapping optical signal sensors \cite{Piacenza2020TMech}, and commercial offerings from SynTouch, Pressure Profile Systems, ATI Industrial Automation, and OnRobot. For recent comprehensive reviews, see \cite{Dahiya2010TRO, Yousef2011SensAct, KapassovRAS2015, Chen2018Sensors, Yamaguchi2019AdvRob}. However, when using such sensors to accomplish low-level or long-horizon control tasks, key challenges remain, including the following:

\begin{itemize} 
\item How does one accurately estimate low-dimensional \textit{tactile features}, such as 3D contact location and net force vector, from raw tactile signals (e.g., electrical or visual output)? These features facilitate classical grasping and manipulation methods, such as assessing grasp stability and planning hand-finger trajectories for manipulating rigid objects \cite{Murray1994Book, Mason2001Book}.

\item How does one estimate high-resolution \textit{tactile fields} (e.g., surface deformations of the tactile sensor) from raw tactile signals? These fields provide high-density information useful for interacting with small, geometrically-irregular, fragile, or compliant objects \cite{Bicchi2000ICRA, Li2001ICRA}.

\item How does one accurately synthesize raw tactile signals from tactile features or fields? Synthesis of such signals can enable training of real-world control policies fully in simulation, a capability that is highly limited for state-of-the-art tactile sensors.
\end{itemize}

This study addresses these questions for a state-of-the-art tactile sensor, the SynTouch BioTac (Figure~\ref{fig:intro}). The BioTac is a commercially-available fingertip-shaped sensor that consists of a flexible rubber skin, an ionically-conductive fluidic layer, and a rigid core. The primary measurement device within the sensor is an array of 19 sensing electrodes and 4 excitation electrodes located on the outer surface of the core \cite{Wettels2008AdvRob, SynTouch2018Manual}. As the BioTac contacts an object, the fluidic layer changes shape, altering the voltages measured at the sensing electrodes ($\mathbb{R}^{19}$); these voltages have a complex and, thus far, undetermined relationship with the surface deformations and distributed forces on the rubber skin. The BioTac was selected due to its high spatial resolution and sensitivity, low stiffness and hysteresis, compact form factor, and widespread use in research.

This paper is an extended version of the work presented by \cite{Narang2020RSS} at the Robotics: Science and Systems (RSS) conference. The shared contributions of both submissions are the following:

\begin{enumerate}
    \item A novel experimental dataset containing 3D contact locations, 3D net force vectors, and BioTac electrode values. The dataset was collected from 9 robot-controlled indenters interacting with 3 BioTacs, with over 400 unique indentation trajectories, 800 total trajectories, and 50k data points after subsampling. Contact location was precisely measured through careful design and calibration of a novel testbed. Approximately 70\% of the trajectories were designed to induce shear forces, a critical mechanical phenomenon in grasping and manipulation. This dataset is hereafter referred to as the \textit{purely experimental} dataset.
    
    \item The first 3D finite-element (FE) model of the BioTac. Although previous studies have hypothesized that ``the deflection of the rubber skin [of the BioTac] is almost impossible to model'' \cite{Ruppel2019IAS}, the FE model captures this behavior and its underlying mechanical phenomena. This model was validated against experimental 3D net force vectors and generalized over the wide range of indenters and indentation trajectories; thus, the model can be used to predict the mechanical behavior of the BioTac in diverse conditions. Correspondingly, a second dataset is provided that contains FE predictions of BioTac surface deformations, aligned with the previously-described experimental data. This dataset is hereafter referred to as the \textit{mixed} dataset.
    
    \item Neural-network mappings from A) electrode signals to tactile features (i.e., experimental 3D contact locations and net force vectors), and B) electrode signals to tactile fields (i.e., FE surface deformations). Mapping A can be used to implement classical grasping and manipulation methods using raw tactile data. Mapping B is a first-of-its-kind, high-density version of Mapping A, which can inform control in far greater detail, as well as motivate the design of novel algorithms to leverage such high-resolution information.
\end{enumerate}

In this extended version, all sections of the paper have been substantially augmented with details, references, and analyses. These additions include an expanded literature review (Section~\ref{sec:related_works}), an overview of FE theory (Section~\ref{sec:methods_fe_theory}), data processing methods for aligning experimental and simulation data (Section~\ref{sec:methods_exp_sim_data_proc}), and an expanded Conclusions section (Section~\ref{sec:conclusions}). Furthermore, this version provides the following novel contributions:

\begin{enumerate}
    \item Analyses and detailed visualizations of the experimental and simulation datasets, including the distributions of contact locations, forces, electrode values, and FE deformations (Section~\ref{sec:methods_exp_data_proc} and Section~\ref{sec:methods_fe_model_config}).
    \item An ablation study that compares how accurately multilayer perceptrons (MLP), 3D voxel-grid-based convolutional neural networks (CNN), and point-cloud-based neural networks \cite{Qi2017NIPS} can regress to tactile features (Section~\ref{sec:results_tactile_features}).
    \item A more detailed analysis of the accuracy of the tactile feature regressions when applied not only to unseen trajectories, but also to unseen indenters and BioTacs (Section~\ref{sec:results_tactile_features}). This analysis elucidates how well our learning-based methods can generalize.
    \item Neural-network mappings \textit{from} tactile fields \textit{to} raw electrode signals (Section~\ref{sec:results_tactile_fields}). These mappings outperform previous efforts to synthesize electrode signals \cite{Ruppel2019IAS, Zapata2020Haptics}.
\end{enumerate} 

Datasets, CAD files for the experimental testbed, FE model files, a short summary video, and the RSS conference presentation video are all available on our project website at \url{https://sites.google.com/nvidia.com/tactiledata}.

\section{Related works} \label{sec:related_works}

This section reviews previous efforts to A) estimate tactile features (specifically, 3D contact location and net force vector) from electrode signals on the BioTac, B) model or estimate tactile fields (specifically, deformation fields or force distributions on the sensor or object) from electrode signals, and C) estimate electrode signals from tactile features or fields.

\subsection{Tactile feature estimation}

\cite{Wettels2009TMech} first estimated force magnitude on an early prototype of the BioTac. The BioTac was attached to a robotic hand, and pinch grasps were performed on a 6-axis force/torque (F/T) sensor to collect ground-truth data. Normal force magnitude was estimated by linear regression, and tangential force magnitude was estimated by a Kalman filter. However, normal and tangential force estimation only used data from 2 and 4 electrodes, respectively.

\cite{Wettels2011Robio} estimated contact location and force vector on a later prototype of the BioTac. The BioTac was manually indented using 4 objects of varying curvature. Ground-truth contact location data was acquired by manually marking target points on the BioTac skin, and ground-truth force data was collected using an F/T sensor. Contact location was estimated using a weighted average of the locations of the 3 electrodes with the highest signals, where the weights were functions of the signal magnitudes. Force vector was regressed using a 2-layer MLP. The study achieved an RMS contact location error of $2.4$-$2.9$~$mm$ and a force magnitude error of $18$-$40\%$. Wettels and Loeb later compared the force estimation network to a Gaussian mixture model, but found the latter to be less accurate \cite{Wettels2014Springer}.

\cite{Su2012FrontInNeurorob} estimated force vector using a calibrated analytical model. Ground-truth data was collected by manually pressing and sliding the BioTac on an F/T sensor. Force was estimated as
\begin{align}
    F_i = \sum_{j=1}^{19} S_i E_j n_{ij}
\end{align}
where $F$ is force, $S$ is a calibrated directional scaling constant, $E$ is the impedance value of an electrode (tared against its resting value), $n$ is the unit normal at an electrode, $i$ is the spatial axis, and $j$ is the electrode number. Summation is written explicitly. Critically, the analytical model assumes that the signal from any electrode provides information about force only in the direction normal to that electrode. Force magnitudes were estimated with an RMS error of $0.42$-$0.48$~$N$ (up to $10\%$) along each spatial axis.

\cite{Lin2013TechRep} estimated contact location and force vector. The BioTac was manually indented using a flat cylindrical indenter, and ground-truth force data was collected using an F/T sensor. Similarly to \cite{Wettels2011Robio}, contact location was estimated using weighted averages of the locations of all the electrodes, where the weights were the absolute values of the signal magnitudes, raised to an exponent to bias towards stronger signals; the computed location was then projected to an approximate geometric model of the BioTac surface. Force vector was estimated using the method from \cite{Su2012FrontInNeurorob}.

\cite{Ciobanu2014CISIS} estimated contact location and force vector. The BioTac was attached to a robotic hand-arm system and contacted cylindrical indenters of various diameters. Ground-truth position data was collected by proprioception on the robotic arm, and force data was collected with a scale. Force magnitude was estimated using support vector regression, achieving a mean error of $0.059$~$N$. However, only small normal forces were measured.

\cite{Su2015Humanoids} estimated force vector using 4 different methods. The BioTac was manually indented by a human fingertip, and ground-truth force data was collected using an F/T sensor. The 4 estimation methods were the analytical method of \cite{Su2012FrontInNeurorob}, locally-weighted projection regression, the 2-layer MLP of \cite{Wettels2011Robio}, and a 4-layer MLP. The $4$-layer MLP performed best, and force magnitudes were estimated with an RMS error of $0.43$-$0.85$~$N$ along each spatial axis.

Most recently, \cite{Sundaralingam2019ICRA} estimated force vector using 3 different methods. Ground-truth force data was collected by 1) manually indenting the BioTac with a large flat object and measuring force using a compliant optical sensor (which experienced significant noise), and 2) attaching the BioTac to a robotic hand and contacting a ball attached to the optical sensor. The preceding interactions were primarily in the normal direction, and contact location was not measured. This ground-truth data was then combined with analytical force estimates from a box-pushing task. The $3$ estimation methods were the analytical model from \cite{Su2012FrontInNeurorob}, the 4-layer MLP from \cite{Su2015Humanoids}, and a 3D voxel-grid-based CNN. The 3D CNN performed best. Force magnitudes were estimated with a median error of $0.32$-$0.51$~$N$ ($35$-$40\%$) on the ground-truth data sources, and force directions were estimated with a median error of $0.07$-$0.39$~$rad$.

In summary, the most comprehensive studies performed manual indentation of a single BioTac, imprecisely measured contact location, and/or applied forces in the normal direction. From the highest-quality datasets, the state-of-the-art in tactile feature estimation from BioTac electrode data is 1) an RMS contact location error of $2$-$3$~$mm$, 2) a median force magnitude error of $0.3$-$0.5$~$N$, and 3) a median force direction error of $0.07$~$rad$. In contrast, an aim of our research was to generate a more diverse, higher-quality dataset and meet or exceed these estimation benchmarks.

\subsection{Tactile field estimation}

No prior studies were found that modeled or estimated deformation fields or force distributions on the surface of the BioTac or a contacting object. The only related effort was by \cite{Wettels2008BioRob}, which modeled the change in cross-sectional area of internal fluidic channels during indentation. A highly simplified, 2D plane-stress FE simulation was constructed for a strip of the BioTac skin, and two channel geometries were qualitatively evaluated. In contrast, an aim of our research was to develop the first accurate 3D FE model of the BioTac, which can be used to compute high-density tactile fields.

\subsection{Electrode signal estimation}

\cite{Ruppel2019IAS} estimated electrode signals from contact location, force vector, and temperature. The BioTac was attached to a robotic hand and contacted a single spherical indenter. Ground-truth contact location was measured using a camera and fiducial tags, with approximate calibration offsets. Force data was collected using an F/T sensor, and temperature was measured using the BioTac itself. Electrode signals were estimated using sequential and branched deep neural networks, with the latter achieving a standardized mean error of $8.6\%$.

\cite{Zapata2020Haptics} estimated electrode signals from depth images, RGB images, and target robotic grasping parameters. BioTac SPs were attached to a robotic hand, and grasps were performed on various real-world objects. 3D object point clouds and RGB images were measured using an RGB-D camera, and target contact points on the object and wrist poses for the hand were generated using a grasp sampler. Electrode signals were estimated using PointNet \cite{Qi2017CVPR}, achieving an RMS error of $5.9$-$6.1\%$ of the working range of the sensor for unseen samples of seen objects, and $8.7$-$10\%$ for unseen objects.

In contrast with the preceding studies, an aim of our research was to estimate electrode signals with lower error from simulated tactile fields, facilitating future simulation-based training of control policies that are accurate and leverage high-density data.

\section{Methods}

\subsection{Experimental testbed design}
\begin{figure}[thpb]
  \centering
  \includegraphics[scale=0.99]{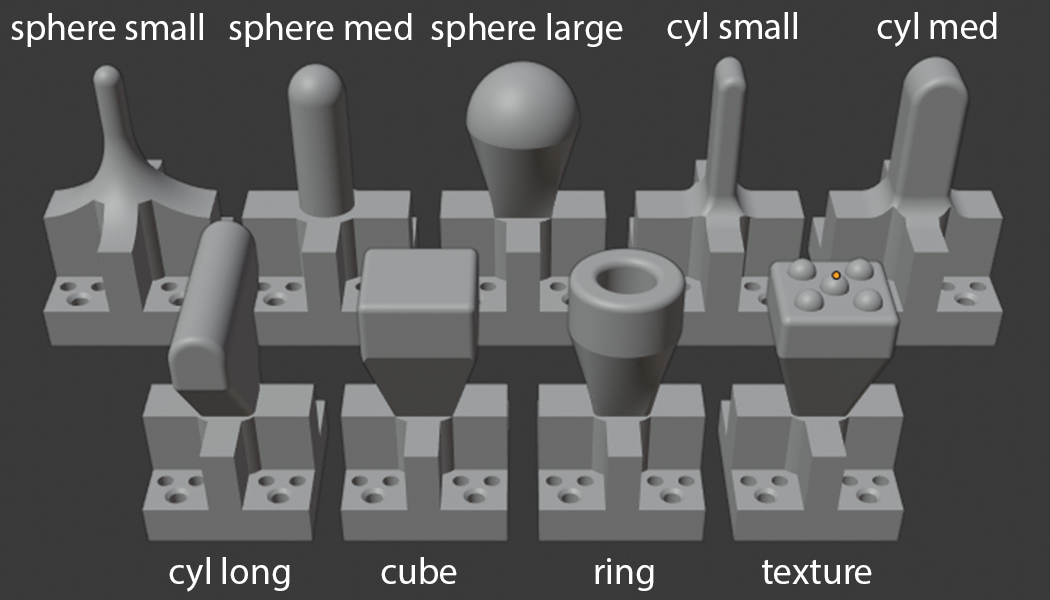}
  \caption{Experimental objects. Nine indenters were designed to capture primitive geometry in everyday objects and surfaces. Example representations include the \textit{cylinder long} indenter, which modeled table and object edges; the \textit{ring} indenter, which modeled bottle openings; and the \textit{texture} indenter, which modeled scattered grains and debris on kitchen surfaces. Critical dimensions of the indenters were scaled to approximately half, equal, or double the radius of the BioTac.}
  \label{fig:methods_indenters}
\end{figure}

To automatically collect high-quality ground-truth 3D contact locations and net force vectors from the BioTac sensor, a custom experimental setup was designed and carefully calibrated. The setup consisted of three major components: 1) an ABB YuMi bimanual robotic manipulator, 2) 3D-printed indenters attached to the distal links of the manipulator, and 3) a mount that rigidly coupled the BioTac to a Weiss Robotics KMS40 6-axis strain-gauge-based F/T sensor. (For videos of the complete setup in action, please see our project website.) The YuMi was selected due to its high positional repeatability ($0.02$~$mm$) and large dexterous workspace achieved by its 2 7-DOF arms. The indenters were designed to capture primitive geometry of everyday household objects (Figure~\ref{fig:methods_indenters}). They were 3D-printed using a Prusa i3 MK3 desktop printer at fine resolution ($0.05$~$mm$) and attached to the distal links of the YuMi using precision alignment pins and metal screws.

\begin{figure}[thpb]
  \centering
  \includegraphics[scale=0.99]{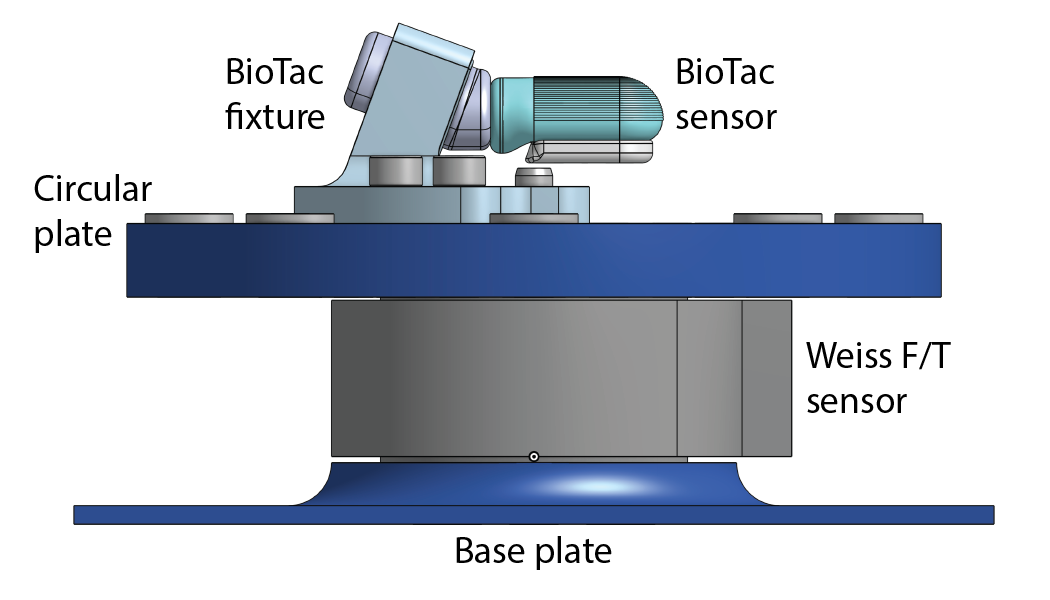}
  \caption{BioTac mount. The mount consisted of 4 parts: a 3D-printed fixture for the BioTac, a 3D-printed circular plate that coupled the BioTac fixture to a Weiss Robotics force/torque sensor, the force/torque sensor itself, and a 3D-printed base plate that allowed the force/torque sensor to be coupled to a benchtop.}
  \label{fig:methods_assembly}
\end{figure}

The mount itself consisted of 4 parts (Figure~\ref{fig:methods_assembly}): 1) a 3D-printed fixture into which the BioTac was inserted, 2) a 3D-printed circular plate that coupled the BioTac fixture to the Weiss F/T sensor, 3) the F/T sensor itself, and 4) a 3D-printed base plate that allowed the F/T sensor to be stably coupled to a benchtop using external C-clamps. The BioTac achieved an interference fit within its fixture and was secured at a predefined depth using a set screw. The fixture, circular plate, F/T sensor, and base plate were all coupled together using precision dowel pins and metal fasteners.

\subsection{Mechanical registration procedure} \label{sec:methods_mech_registration}

\begin{figure}[thpb]
  \centering
  \includegraphics[scale=0.23]{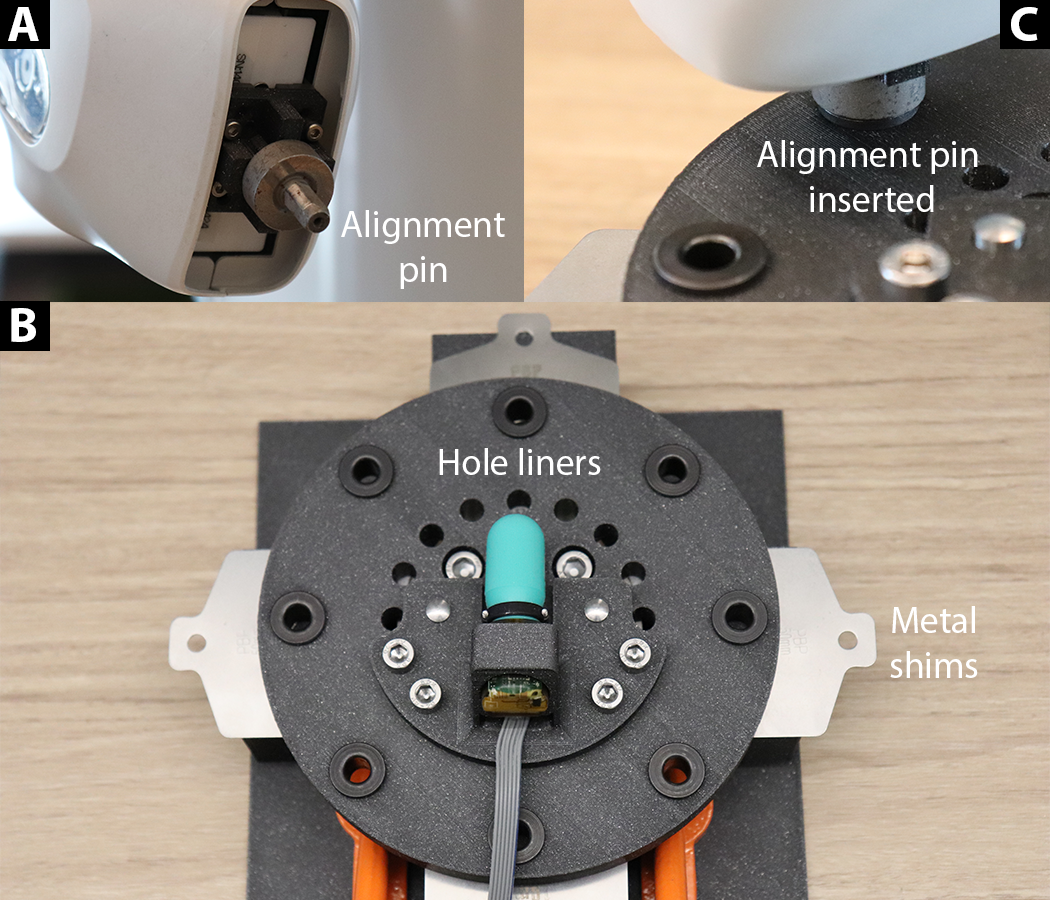}
  \caption{Mechanical registration apparatus. A) A precision alignment pin attached to the distal link of one of the arms of a YuMi robot. B) An overhead view of the apparatus, illustrating precision metal hole liners and temporarily-inserted metal shims. C) The alignment pin inserted into a hole liner during the registration procedure.}
  \label{fig:methods_setup}
\end{figure}

To spatially register the 6D pose of the robot end-effectors with respect to the BioTac, a mechanical registration apparatus inspired by computer numerical control (CNC) alignment techniques was first designed and constructed. The apparatus consisted of 2 major components: diamond-head shoulder-style metal alignment pins attached to the distal links of the 2 arms of the YuMi (Figure~\ref{fig:methods_setup}A), and metal hole liners that were press-fit into corresponding holes along the circumference of the circular plate (Figure~\ref{fig:methods_setup}B). The worst-case clearance between the alignment pins and hole liners was approximately $0.3$~$mm$. Metal shims were temporarily inserted between the circular plate and the F/T sensor to prevent large loads during registration from pitching the plate. They were removed following registration.

During registration, lead-through mode on the YuMi was enabled, and the alignment pin for a given arm was manually guided into a metal hole liner until the bottom face of the pin's shoulder rested flush against the top face of the hole liner (Figure~\ref{fig:methods_setup}C). Upon contact, the end-effector position was recorded in a robot-fixed coordinate frame $R$ as vector $\textbf{p}_1^R$ (using a forward kinematic model of the YuMi and alignment pin) and in a BioTac-fixed coordinate frame $B$ as vector $\textbf{q}_1^B$ (using a CAD model of the undeformed BioTac and experimental mount). This step was repeated for 2 additional hole liners, resulting in measurements $p^R = \{\textbf{p}_1^R, \textbf{p}_2^R, \textbf{p}_3^R\}$ and $q^B = \{\textbf{q}_1^B, \textbf{q}_2^B, \textbf{q}_3^B\}$. Note that $p^R$ and $q^B$ correspond to 2 physically distinct sets of non-collinear position vectors, where the endpoints of a given $\textbf{p}_i^R$ and $\textbf{q}_i^B$ are coincident.

Following the algorithm proposed by \cite{Marvel2016ISAM}, vectors $\textbf{v}_1^R = \textbf{p}_2^R - \textbf{p}_1^R$ and $\textbf{v}_2^R = \textbf{p}_3^R - \textbf{p}_1^R$ were computed for $R$, and vectors $\textbf{v}_1^B = \textbf{p}_2^B - \textbf{p}_1^B$ and $\textbf{v}_2^B = \textbf{p}_3^B - \textbf{p}_1^B$ were computed for $B$. Note that $v^R = \{\textbf{v}_1^R, \textbf{v}_2^R\}$ and $v^B = \{\textbf{v}_1^B, \textbf{v}_2^B\}$ correspond to a physically identical set of 2 non-orthonormal basis vectors for a common intermediate frame $I$, expressed in $R$ and $B$, respectively. Next, using sequences of cross products, orthonormal basis vectors $x^R = \{\hat{\textbf{x}}^R, \hat{\textbf{y}}^R, \hat{\textbf{z}}^R\}$ were generated for $I$ from $v^R$, and  physically identical vectors $x^B = \{\hat{\textbf{x}}^B, \hat{\textbf{y}}^B, \hat{\textbf{z}}^B\}$ were generated from $v^B$. Finally, $x^R$ and $x^B$ were used to define homogeneous transformation matrices ${}^R_I H$ and ${}^B_I H$, respectively; the desired transformation matrix from the robot-fixed coordinate frame to the BioTac-fixed coordinate frame was simply 
\begin{align}
    {}^B_R H = {}^B_I H ({}^R_I H)^{-1}
\end{align}

\begin{figure}[thpb]
  \centering
  \includegraphics[scale=0.23]{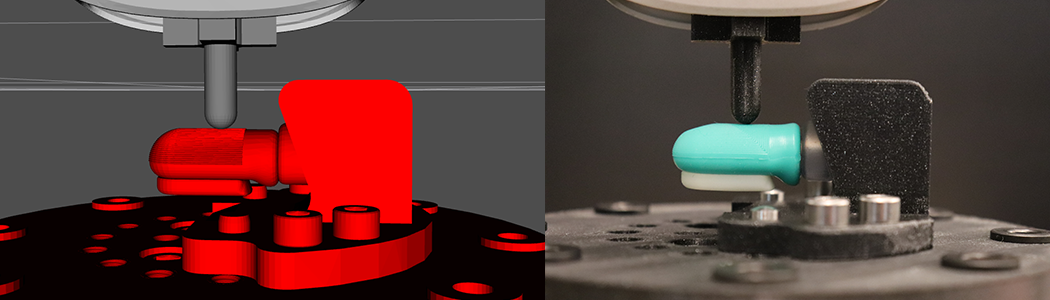}
  \caption{Example of registration accuracy. Left: Predicted poses of the \textit{sphere medium} indenter and the BioTac, with the indenter just initiating contact with the rubber skin. Right: Observed poses.}
  \label{fig:methods_registration}
\end{figure}

To ensure robustness to measurement error, the previous procedure was repeated for all 3-hole combinations of 4 different hole liners on the circular plate. Following the method proposed and validated by \cite{VanWyk2018TASE}, the transformation matrices for all combinations were then averaged to produce a final transformation matrix. Registration error between the robot and BioTac was observed to be between $0.5$-$1.5$~$mm$ (Figure~\ref{fig:methods_registration}) over the full range of robot joint configurations and end-effector positions traversed during testing. The full mechanical registration procedure was applied to the right and left arms of the YuMi independently.

\subsection{Experimental testing procedure} \label{sec:methods_exp_testing_procedure}

\begin{figure}[thpb]
  \centering
  \includegraphics[scale=0.99]{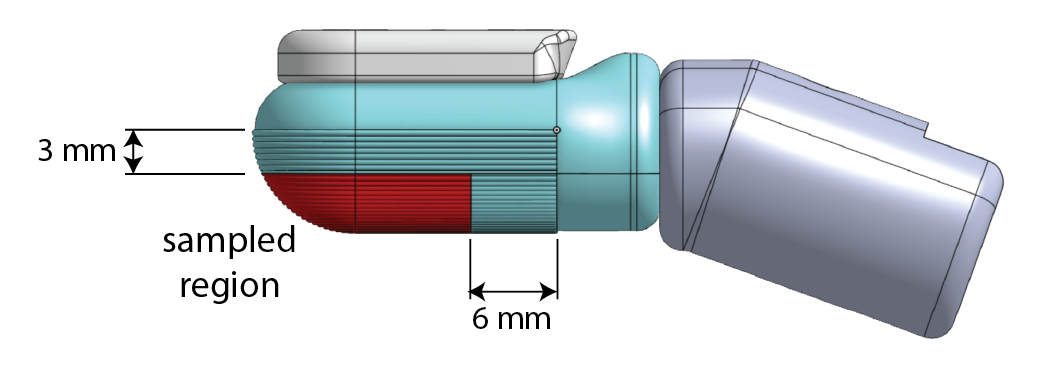}
  \caption{Point sampling. Target points were sampled from the indicated region, with dimensions defined with respect to the bounds of the BioTac fingerprints. Due to the width of the indenters and the angular orientations of the trajectories, contact frequently occurred well outside the sampled region.}
  \label{fig:methods_sampling}
\end{figure}

For each of the 9 indenters (Figure~\ref{fig:methods_indenters}), 10 unique points were randomly sampled from a subsection of the ventral surface of the BioTac (Figure~\ref{fig:methods_sampling}), referred to here as the \textit{sampled region}. With respect to the long axis of the BioTac, points on the side closer to the right arm of the YuMi were assigned to that arm for indentation, and vice versa for points closer to the left arm. For each point, an indentation trajectory was generated normal to the surface. In addition, 4 angled trajectories were generated along lines oriented at $30$~$deg$ from the surface normal in order to purposely induce shear loading, which has largely been avoided or mitigated in previous studies. Note that due to the width of the indenters, as well as the angles of the trajectories, contact with the BioTac frequently occurred well outside the sampled region.

Each trajectory was designed to begin $3$~$mm$ away from the BioTac surface and ultimately indent the sensor by $3$~$mm$ (for normal indentations) or $1.5$~$mm$ (for angled indentations) past the point of initial contact. In addition, each trajectory was divided into $0.1$~$mm$ displacement increments, with $5$~$sec$ allotted to each increment in order to allow fluid-structure interaction dynamics in the BioTac to settle. All trajectories and joint commands were generated using Riemannian Motion Policies (RMP) \cite{Ratliff2018ArXiv}, which are well suited to position-control applications requiring high accuracy, avoidance of self-collisions, and avoidance of specific obstacles. Trajectories were simulated using the ROS \textit{rviz} package, and specific trajectories were eliminated that could not be achieved with sufficiently high end-effector accuracy (i.e., $0.01$~$mm$ or better, according to a forward kinematic model of the YuMi and indenter) or resulted in unexpected collisions with the BioTac mount.

Prior to execution of the trajectories on the real-world experimental setup, the BioTac was connected to power and left untouched for approximately $30$~$min$. The conductivity of the BioTac fluid and the corresponding electrode values are dependent on temperature \cite{Wettels2008AdvRob, Ruppel2019IAS}; this waiting period allowed the BioTac to reach thermal equilibrium. All trajectories were then executed, and the experiments were repeated for $3$ different BioTacs. However, 1 sensor (i.e., BioTac~1) had data exclusively from the \textit{sphere small}, \textit{sphere medium}, and \textit{sphere large} indenters due to accidental damage during testing. Joint angles from the active YuMi arm ($\mathbb{R}^{7}$), net force/torque data from the Weiss F/T sensor ($\mathbb{R}^{6}$), and electrode values from the BioTac ($\mathbb{R}^{19}$) were continuously acquired at a minimum of $100$~$Hz$ using ROS. Joint angles were converted to indenter tip positions ($\mathbb{R}^{3}$) using a forward kinematic model of the YuMi and indenter; note that in this paper, \textit{3D contact locations} are defined as these tip positions after contact with the BioTac. Electrode values were normalized from a raw $12$-$bit$ range to a $[0, 1]$ interval. The tip positions, force/torque data, and electrode values were all time-stamped and recorded. For later subsampling (see Section~\ref{sec:methods_exp_data_proc}), time stamps were also separately recorded at the end of each $0.1$~$mm$ displacement increment within each trajectory.

\subsection{Experimental data processing} \label{sec:methods_exp_data_proc}

The following post-processing steps were performed on the experimental data to produce the \textit{purely experimental} dataset:
\begin{enumerate}
    \item To reduce the dataset size ($>1.5e6$ samples) and mitigate redundancy, the indenter tip positions, force/torque data, and electrode values were subsampled at the final time step of each $0.1$~$mm$ displacement increment within each trajectory.
    \item To mitigate drift, force/torque values and BioTac electrode values for each indentation were tared against their respective values at the first time step of the trajectory (i.e., before contact between the BioTac and indenter had occurred).
    \item Due to noise on the F/T sensor readings, a 1st-order Butterworth low-pass filter with a cutoff frequency of $5$~$Hz$ was applied forward and backward to the force data. However, at net force magnitudes below $0.5$~$N$, substantial noise persisted. Thus, experimental data corresponding to forces below this threshold were filtered out; note that previous studies have set this value higher (e.g., $1.0$~$N$ in \cite{Lin2009Robio}. After filtering, initial experimental data points corresponded to force magnitudes slightly higher than $0.5$~$N$; to facilitate later alignment with simulation data (see Section~\ref{sec:methods_exp_sim_data_proc}), linear interpolation was used to prepend experimental data points corresponding to \textit{exactly} $0.5$~$N$.
\end{enumerate}

\begin{figure}[thpb]
  \centering
  \includegraphics[scale=0.99]{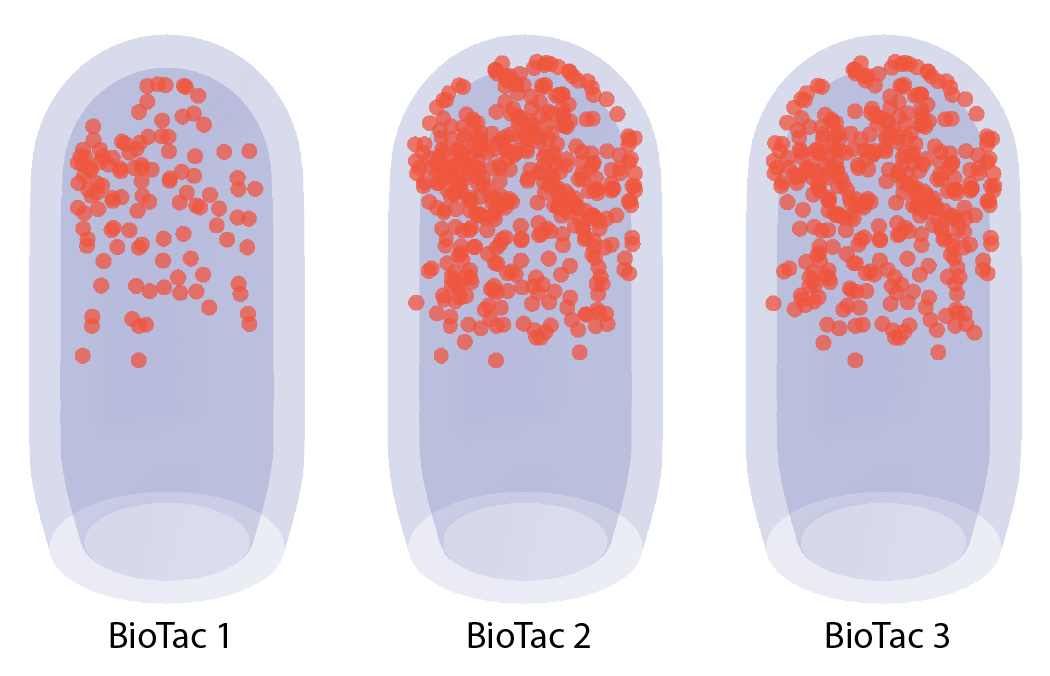}
  \caption{Experimental contact locations. The inner and outer ventral surfaces of the undeformed BioTac skin are shown in translucent blue. Red markers indicate contact locations and are given for the final displacement increment of each trajectory. Note that BioTac~1 had fewer locations, as only 3 indenters were used (see Section~\ref{sec:methods_exp_testing_procedure}).}
  \label{fig:data_contact}
\end{figure}

\begin{figure}[thpb]
  \centering
  \includegraphics[scale=0.99]{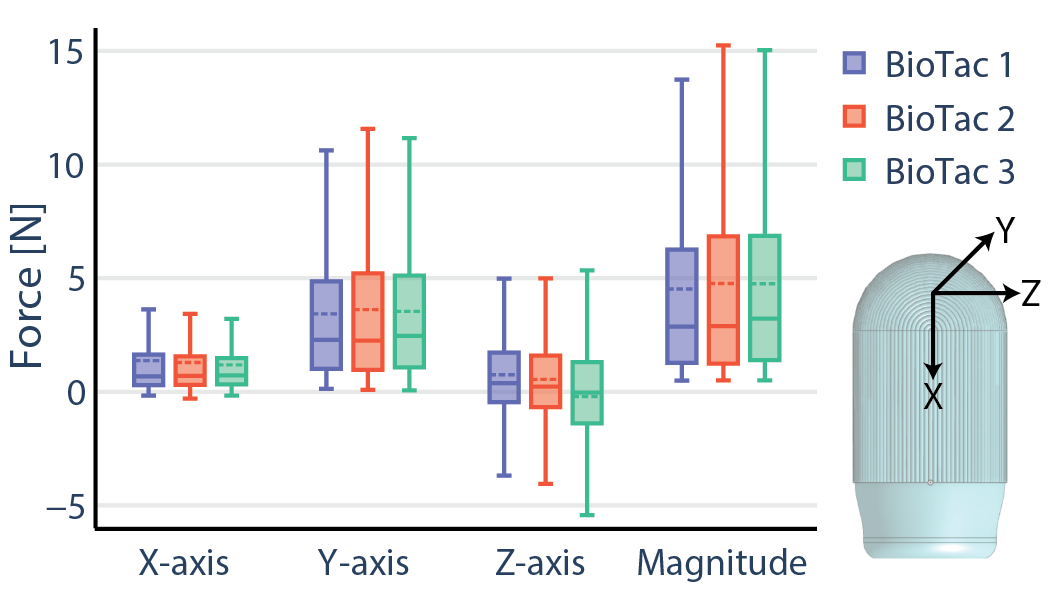}
  \caption{Experimental force values. The solid and dashed horizontal lines in each box indicate the median and mean, respectively. The box height is equal to the interquartile range (IQR). The fence length is equal to $1.5 * IQR$, rounded down to the nearest data point. The maximum force magnitudes (not pictured) for BioTac~1, 2, and 3 were $23.4$, $32.1$, and $29.0~N$, respectively.}
  \label{fig:data_force}
\end{figure}

\begin{figure*}[thpb]
  \centering
  \includegraphics[scale=1.01]{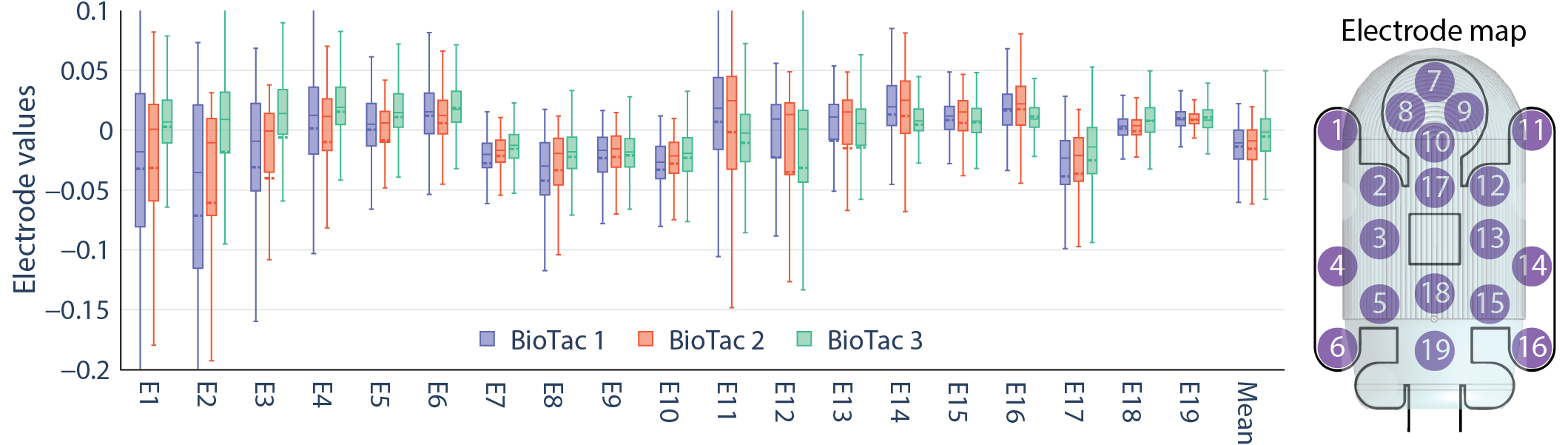}
  \caption{Experimental electrode values. Left: Box plots for all BioTacs and electrodes. Electrodes are denoted by the $E$ prefix on the $x$-axis, and electrode values are normalized and tared as described in Section~\ref{sec:methods_exp_testing_procedure} and Section~\ref{sec:methods_exp_data_proc}. The solid and dashed horizontal lines in the boxes indicate the median and mean, respectively. The box height is equal to the interquartile range (IQR). The fence length is equal to $1.5 * IQR$, rounded down to the nearest data point. The full range of electrode values (not pictured) for BioTac~1, 2, and 3 were $[-0.65, 0.13]$, $[-0.78, 0.11]$, and $[-0.51, 0.17]$, respectively, where the theoretical maximum span was $1$. Note that higher IQR for certain electrodes does not necessarily indicate that greater or more frequent deformation was experienced near those electrodes, as electrode value has a highly nonlinear relationship with deformation that varies from electrode to electrode. Right: Map of electrodes on the BioTac. An unwrapped electrode map is shown superimposed on a rounded ventral view of the BioTac.}
  \label{fig:data_electrode}
\end{figure*}

After post-processing, the dataset contained 411 unique trajectories, 889 total trajectories, 17703 time steps, and 53109 data points. Figure~\ref{fig:data_contact}, Figure~\ref{fig:data_force}, and Figure~\ref{fig:data_electrode} illustrate the distributions of the 3D contact locations, 3D force vectors, and electrode values. As depicted, a diverse set of interactions was explored, and a broad range of sensor values was collected. One subtlety is that the contact locations and force values were highly consistent from BioTac to BioTac, whereas electrode values were highly inconsistent; this lack of consistency likely reflects electronic manufacturing variability within the BioTacs, complicating the task of transfer learning from one device to another. (See Section~\ref{sec:results_tactile_features} for further details.)

\subsection{Finite element theory} \label{sec:methods_fe_theory}

For tactile fields, such as surface deformations on the BioTac and distributed forces transmitted across the BioTac-object interface, real-world ground-truth data is prohibitively difficult to acquire. To generate near-ground-truth data for such quantities, a finite element (FE) model was developed. Given the importance of the FE method to the present work, as well as other robotics studies involving deformable-object simulation, a brief overview is provided here.

The FE method is a well-established numerical technique for solving partial differential equations (PDEs). As an example relevant to the present work, the PDEs for the deformation of a 3D solid object that 1) is in static equilibrium, 2) has a linear stress-strain relation, and 3) experiences small deformations, can be written concisely as
\begin{align}
    \sigma_{ji,j} + f_i = 0 \label{eq:static_equilibrium} \\
    \sigma_{ij} = C_{ijkl} \epsilon_{kl} \label{eq:hookes_law} \\
    \epsilon_{ij} = \frac{1}{2} (u_{i,j} + u_{j,i}) \label{eq:strain_displacement}
\end{align}
where Equation~\ref{eq:static_equilibrium} expresses the static equilibrium equations ($\sigma$ is the 2nd-order stress tensor; $f$ is the body force per volume; $i$, $j$, $k$, and $l$ denote spatial axes; the comma denotes a partial derivative; and Einstein summation convention is used); Equation~\ref{eq:hookes_law} expresses the linear stress-strain equations, also known as Hooke's Law ($\epsilon$ is the 2nd-order strain tensor, and $C$ is the 4th-order stiffness tensor that relates $\sigma$ to $\epsilon$); and Equation~\ref{eq:strain_displacement} expresses the small-deformation strain-displacement equations ($u$ is displacement). These systems of PDEs can often be condensed into a single governing PDE for particular problems (e.g., bending beams). Additional equations are added to apply boundary conditions (i.e., geometric constraints or loading conditions).

In general, such PDEs are challenging to analytically solve. For the 3D deformable solid, solving a governing PDE to compute a displacement field $u_i$ along a spatial axis $i$ is only tractable for the simplest boundary conditions. A basic numerical approach would attempt to approximate the solution as a linear combination of functions, such as
\begin{align}
    u_i = \sum_m^N c_m \phi_m \label{eq:linear_combination}
\end{align}
where $\phi_m$ are \textit{basis functions} to be chosen (e.g., polynomials); $c_m$ are unknown coefficients to be solved for; $m$ is not a spatial axis, but an arbitrary index; and $N$ is a finite positive integer, which determines the number of points at which the PDE is to be satisfied. However, for a given PDE, boundary conditions, $\phi_m$, and solution points, there may be no exact solution for the $c_m$; furthermore, even exact solutions may result in $u_i$ that are inaccurate over most of the spatial domain.

To address these issues, a weighted-integral formulation may be used, such as
\begin{align}
    \int_{V} w_m R(u_i) dV = 0
\end{align}
where $R$ is the residual of a PDE (i.e., a simple rearrangement), $w_m$ are \textit{weight functions} to be chosen (e.g., could be polynomials as well), and $V$ is volume. The approximation for the solution to the PDE (e.g., Equation~\ref{eq:linear_combination}) can be substituted into $R$, and the $c_m$ can be solved for once again. Intuitively, the PDE is now satisfied on average over the entire spatial domain, according to multiple weight distributions determined by the $w_m$. However, such integrals may be difficult to evaluate; furthermore, for an arbitrary problem with complex geometry and boundary conditions, determining reasonable $\phi_m$ and $w_m$ may be challenging.

The FE method approaches this issue by dividing a complex geometrical domain (e.g., the BioTac) into simple geometric subregions. The subregions are typically referred to as \textit{elements}, and the vertices of the elements as \textit{nodes}; all elements and nodes collectively define a \textit{mesh}. For 3D geometries, the elements often have a tetrahedral or hexahedral shape. The weighted-integral equations are now formulated for each element (specifically, the \textit{weak form} of these equations, which reduces differentiability requirements for the $\phi_m$). Typically, $N$ is assigned to be the number of degrees of freedom of the element along a spatial axis (i.e., the number of nodes); the $c_m$ are assigned to be the unknown nodal positions $u_m$; the $\phi_m$ are assigned to be polynomials that ensure that $u_i = u_m$ at each node; and the $w_m$ are chosen to be identical to the $\phi_m$.

As displacements are continuous from element to element, the weighted-integral equations cannot be solved for each element individually. Instead, a system of coupled equations for all elements is assembled, which has form
\begin{align}
    K_{pq} u_q + C_{pq} \frac{\partial u_q}{\partial t} = M_{pq} \frac{\partial^2 u_q}{\partial t^2}
\end{align}
where $K$, $C$, and $M$ are referred to as the stiffness, damping, and mass matrices, respectively; $u$ is the vector of all nodal positions; and $p$ and $q$ denote spatial axes. This system of equations can be solved via numerical linear algebra techniques. In essence, the FE method transforms PDEs into large sets of algebraic equations that resemble mass-spring-damper equations, with a systematic approach for determining the matrix coefficients. With high mesh density and sufficiently small time steps, predictions for the deformation of solids can be extremely accurate \cite{Reddy2019Book}, even for highly nontraditional mechanical systems \cite{Narang2018AFM}.

\subsection{Finite element model configuration} \label{sec:methods_fe_model_config}

In this work, the FE model was developed and simulated using ANSYS Mechanical v19.1, an industry-standard nonlinear FE software. To match the quasistatic conditions of the real-world experiments, static structural simulations were performed.

\begin{figure}[thpb]
  \centering
  \includegraphics[scale=0.23]{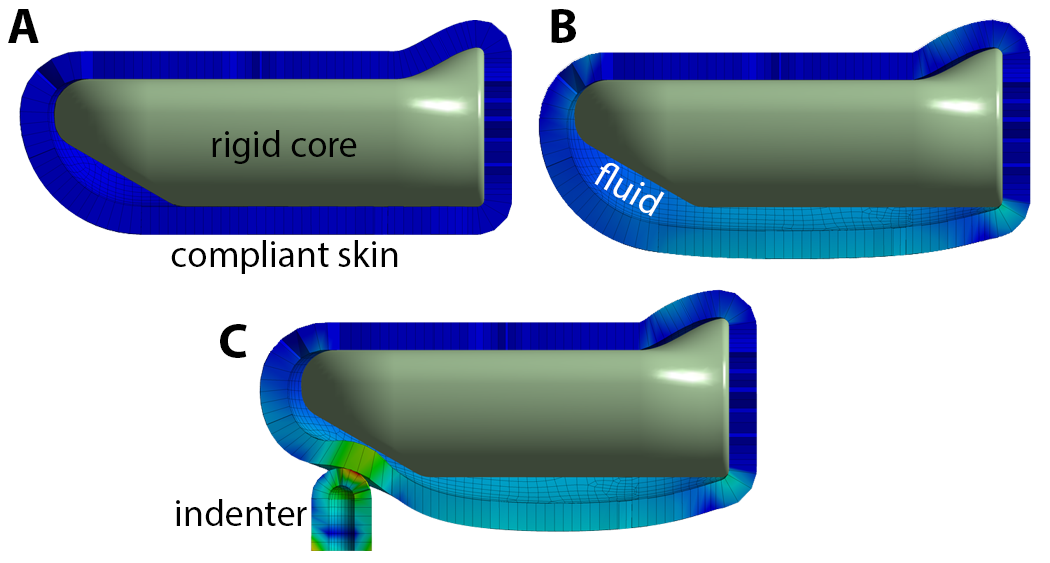}
  \caption{Finite element model deformation. A) A cross-sectional view during an example simulation, prior to load step 1 (i.e., increase of volume of the fluid). B) The final time step of load step 1. C) A time step near the end of load step 2 (i.e., indentation). Color gradients indicate von Mises equivalent stress.}
  \label{fig:methods_ansys}
\end{figure}

Geometrically, the FE model consisted of 4 components: 1) the BioTac skin, 2) the BioTac core, 3) the fluid between the skin and the core, and 4) the indenter in contact with the BioTac (Figure~\ref{fig:methods_ansys}). The initial CAD models of the skin and core were acquired from the manufacturer; however, the asperities on the interior and exterior surface of the skin (e.g., fingerprints) were removed to facilitate meshing and expedite convergence. Four-node shell elements (SHELL181) were selected for the skin due to their high efficiency and accuracy for modeling large, nonlinear deformation of thin and thick membranes \cite{Lee2018Book}. The core was defined as a rigid body, and hydrostatic fluid elements (HSFLD242) were used for the fluid layer to efficiently enforce physically-realistic incompressibility (i.e., conservation of the fluidic volume). Finally, rather than defining the indenter as a rigid body, 4-node shell elements were used to enable extraction of distributed loads. Large-deformation analysis was enabled, allowing the strain-displacement relations of Equation~\ref{eq:strain_displacement} to be complemented with terms that are quadratic in the partial derivatives of $u$.

The stress-strain relations for the BioTac skin were then defined, as well as the material properties of the fluid. The BioTac skin is made of silicone rubber; the stress-strain behavior of such a material is highly nonlinear, and consequently, cannot be accurately modeled with linear relations, such as those of Equation~\ref{eq:hookes_law}. Instead, rubber is typically modeled with a \textit{hyperelastic} material model, in which the stress-strain relationship is nonlinear, but energy is conserved during loading.

For mathematical convenience, the stress-strain relations for a hyperelastic material are often defined indirectly in terms of an \textit{energy density function} $W(C_{ij})$, where $C$ is the \textit{right Cauchy-Green deformation tensor}, a deformation metric that describes how the squares of distances change during deformation; and $i$ and $j$ denote spatial axes. Note that the deformation at any point in the material can always be reexpressed as extensions or compressions along 3 principal axes. As energy density does not change with the coordinate frame, $W$ can be rewritten as a function of the tensor invariants of $C_{ij}$, which themselves are functions of the principal stretches $\lambda_1$, $\lambda_2$, and $\lambda_3$; these $\lambda_k$ are strain-like quantities that describe the length of the principal axes after deformation, normalized by original length. Finally, from the work-energy principle, the stress-strain relations along the principal axes can be derived as
\begin{align}
    \sigma_l = \frac{1}{\lambda_m \lambda_n} \frac{\partial W}{\partial \lambda_l}
\end{align}
where $l$, $m$, and $n$ denote arbitrary principal axes, with $l \neq m \neq n$.

Due to its mathematical simplicity and accuracy, the \textit{incompressible Neo-Hookean} energy density function was chosen for the skin. This function can be written as
\begin{align}
    W = \frac{\mu_{NH}}{2}(\lambda_1^2 + \lambda_2^2 + \lambda_3^2) \label{eq:neo_hookean}
\end{align}
where $\mu_{NH}$ is a specified stiffness parameter (assigned in Section~\ref{sec:methods_fe_model_calibration} via optimization), and stretches $\lambda_i$ satisfy the incompressibility constraint $\lambda_1 \lambda_2 \lambda_3 = 1~($\cite{Treloar1974Book}. As for the fluid, although the BioTac may be approximated as isothermal after an initial waiting period, an arbitrary volumetric thermal expansion coefficient $\beta_T = 0.01$~$C^{-1}$ was defined as a simple means to later adjust the fluidic volume (as described later in this section).

From simulation and physical considerations, three spatial boundary conditions were then defined: 1) grounding of the rigid core, which removed rigid-body modes from the simulation, 2) anchoring together the proximal (i.e., right, in Figure~\ref{fig:methods_ansys}) surfaces of the skin and core, which modeled the effect of the circumferential BioTac clamp, and 3) anchoring together the dorsal (i.e., top, in Figure~\ref{fig:methods_ansys}) surfaces of the skin and core, which modeled the effect of the BioTac nail. In addition, 2 contact boundary conditions were defined: frictionless contact between the skin and core (due to lubrication provided by the fluid), and frictional contact between the indenter and the skin. The normal-Lagrange contact formulation was selected to minimize interpenetration. A mesh refinement study was used to prescribe a maximum mesh dimension of $0.5$~$mm$; less than this value, changes in simulation results were negligible.

Two load steps were then applied: 1) a pressurization step, in which the volume of fluid between the skin and core was gradually increased over 5 equal time steps (Figure~\ref{fig:methods_ansys}B), and 2) an indentation step, in which displacements were applied to the indenter over 20 equal time steps (Figure~\ref{fig:methods_ansys}C). The pressurization step was necessary because the manufacturer only provided CAD models of the unpressurized BioTac (i.e., no fluid); in simulation, the temperature of the fluid was increased until the total sensor thickness matched the manufacturer's specification of $15.1$~$mm$ \cite{SynTouch2018Manual}. For step 2, the indentation trajectories were specified to exactly match the target trajectories in the real-world experiments. An asymmetric Newton-Raphson solver was selected to facilitate convergence for frictional contact between the indenter and the skin.

\begin{figure}[thpb]
  \centering
  \includegraphics[scale=0.99]{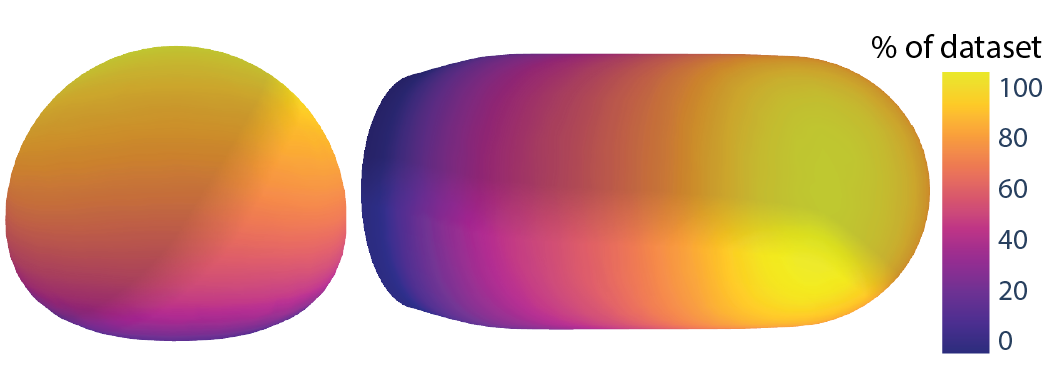}
  \caption{Finite element mesh motion. A front view and ventral view of the undeformed BioTac mesh are shown. For each point on the mesh, the color indicates how frequently that point exhibited significant motion, expressed as a percentage of all the time steps in the dataset. Here, \textit{significant motion} is defined as a displacement magnitude greater than $0.1$~$mm$.}
  \label{fig:data_mesh}
\end{figure}

The simulation was executed using 6 CPUs. Two primary quantities were extracted: 1) \textit{nodal coordinates} at the end of step 1, which served as reference coordinates, and 2) \textit{nodal displacements} throughout step 2, relative to the reference coordinates. For later validation, the preceding simulation and data extraction procedure was repeated for every indenter and target trajectory examined in the real-world experiments, taking an average of $7$~$min$ per trajectory. Figure~\ref{fig:data_mesh} illustrates the frequency of motion of each point on the BioTac mesh over all simulations. The broad distribution reflects the high diversity of interactions.

\subsection{Experimental and simulation data processing} \label{sec:methods_exp_sim_data_proc}

In Section~\ref{sec:methods_fe_model_calibration}, Section~\ref{sec:methods_fe_model_validation} and Section~\ref{sec:methods_tactile_fields}, experimental and simulation data will be compared and combined for FE model calibration, FE model validation, and estimation of tactile fields. However, experimental and simulation data exhibited structural differences, and post-processing was required to ensure that the data could be appropriately compared. The following steps were performed to produce the \textit{mixed} dataset (i.e., simulations aligned with experiments):

\begin{enumerate}
    \item All trajectories generated in Section~\ref{sec:methods_exp_testing_procedure} were executed in simulation. However, as noted in the same section, some of these trajectories were disqualified for experimental testing due to anticipated inaccuracy or collisions. Furthermore, a very small number of trajectories failed in the real-world due to spurious sensor saturation. All trajectories common to both experiment and simulation were compared.
    \item For consistency with the experimental data (see Section~\ref{sec:methods_exp_data_proc}), simulation data corresponding to net force magnitudes below $0.5$~$N$ were filtered out, and linear interpolation was used to prepend simulation data points corresponding to \textit{exactly} $0.5$~$N$. Thus, the starting points of the experimental and simulation datasets were exactly aligned according to force.
    \item During contact between the skin and the internal rigid core, simulations often diverged due to the high stiffness of the incompressible rubber, combined with the moderately-large time steps used to ensure a reasonable simulation time. Simulation data at or following these divergence points were filtered out.
    \item For a given trajectory, starting at the data points corresponding to $0.5$~$N$, the total experimental and simulation indenter displacements were often slightly different, primarily due to simulation divergence. The minimum final displacement of the experimental and simulation trajectories was determined, and experimental and simulation data beyond that point were disregarded.
    \item After the preceding steps were completed, experimental and simulation data were both linearly interpolated along 20 evenly-spaced indenter displacement increments. Thus, the datasets could now be readily compared.
\end{enumerate}

\subsection{Finite element model calibration} \label{sec:methods_fe_model_calibration}

As FE models express the laws of physics, they are highly constrained in comparison to data-driven approaches to predicting physical systems. However, a small number of free parameters still need to be measured or estimated. The current model has 4 parameters: 1) the thickness of the skin $t_{sk}$, which was geometrically modified to remove surface asperities (and thus, requires tuning to achieve accurate deformation relative to the original CAD model), 2) the coefficient $\mu_{NH}$ in the Neo-Hookean energy density function (Equation~\ref{eq:neo_hookean}), 3) the coefficient of friction $\mu_{fr}$ between the indenter and the skin, and 4) the temperature of the fluid $T_{fl}$ that, given parameters 1 and 2, achieves a desired sensor thickness of $15.1$~$mm$ at the end of the first load step.

\begin{figure}[thpb]
  \centering
  \includegraphics[scale=1.0]{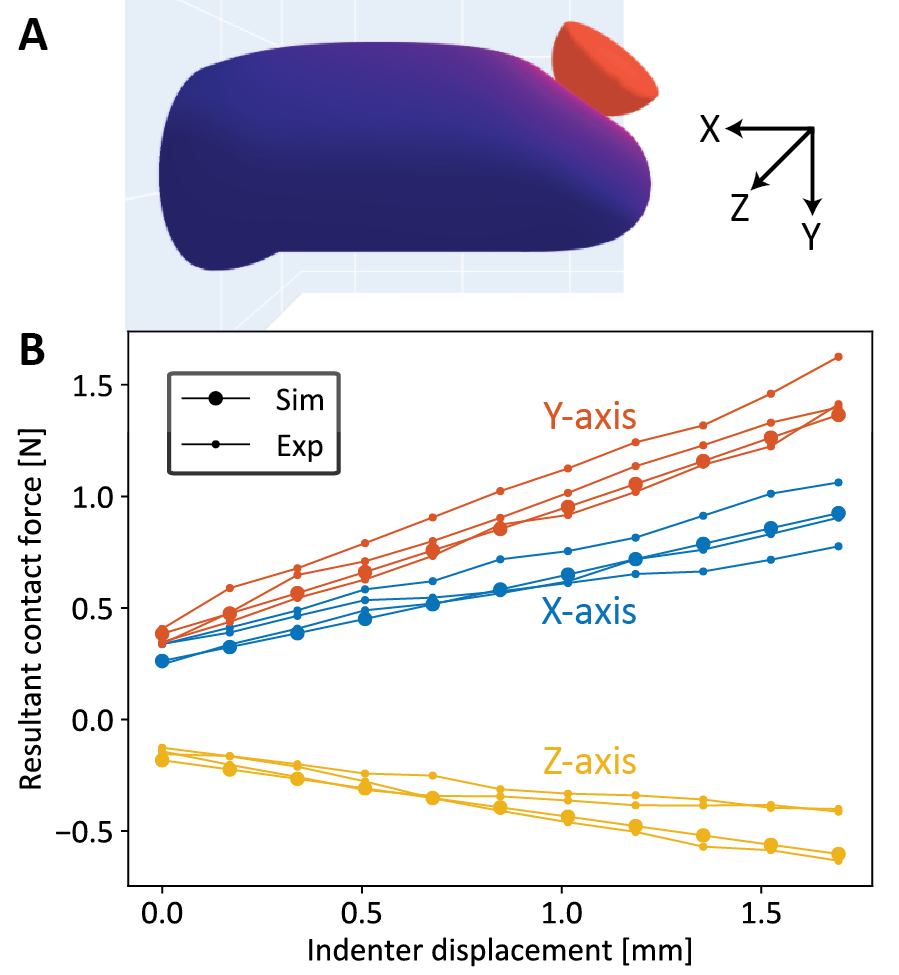}
  \caption{Finite element model calibration. A) The angled trajectory used for the calibration of finite element parameters. B) The corresponding simulation and experimental data for the net contact force acting on the BioTac along each spatial axis. Simulation data after optimization is shown. Experimental data is given for 3 different BioTacs.}
  \label{fig:methods_optim_calib}
\end{figure}

To calibrate the values of these parameters, the net contact force on the BioTac was compared between FE simulations and corresponding experiments, and the parameters were tuned to minimize the mean squared error (MSE) over all time steps. Due to the small number of parameters, it was hypothesized that minimizing force error for a \textit{single} indentation trajectory would produce parameters that generalize well across \textit{all} indentations. As follows, a specific angled trajectory of the \textit{sphere medium} indenter was selected due to its moderate contact area, localized deformation, and generation of both normal and shear stresses (Figure~\ref{fig:methods_optim_calib}); future work may leverage data from multiple trajectories for even further accuracy.

A sequential least-squares programming (SLSQP) optimizer was used to minimize the following cost function:
\begin{align}
    J_{total} & = w_1 J_{force} + w_2 J_{thick} \label{eq:cost_function} \\
    & = w_1 \mathlarger{\sum_{i=1}^3} \sqrt{\frac{1}{N} \sum_{j=1}^N (F_{ij}^{sim}-F_{ij}^{exp})^2} \nonumber \\
    & + w_2(t_{sn}^{sim}-0.0151) \nonumber
\end{align}
where $w_1$ and $w_2$ are hand-selected weights; $J_{force}$ is the sum of the RMS contact force error for each spatial axis; $J_{thick}$ is a cost function ensuring that the total sensor thickness after pressurization matches the manufacturer's specification; $N$ is the total number of samples (i.e., product of number of sensors and time steps per sensor); $F_{ij}^{sim}$ and $F_{ij}^{exp}$ are the simulated and experimental force, respectively, for spatial axis $i$ and sample number $j$; $t_{sn}^{sim}$ is the simulated total sensor thickness; and $0.0151$~$m$ is the manufacturer-specified thickness. Note that $F_{ij}^{sim}$ and $t_{sn}^{sim}$ are functions of simulation parameters $t_s$, $\mu_{NH}$, $\mu_{fr}$, and $T_{fl}$.

Weights $w_1$ and $w_2$ were selected to be $1$ and $1e4$, respectively, such that $J_{force}$ was approximately 1 order of magnitude higher than $J_{thick}$ throughout optimization. From physical consideration of the components comprising the BioTac, upper and lower bounds on the simulation parameters were chosen to represent large, but realistic ranges. The bounds, as well as the optimized values of the parameters, are given in Table~\ref{tab:methods_optim}. FE predictions and experimental values for the net force after optimization are given in Figure~\ref{fig:methods_optim_calib}.

\begin{table}[thpb]
\small\sf\centering
\caption{Optimization parameters. Upper and lower bounds for the parameters are given, as well as optimized values.\label{tab:methods_optim}}
\begin{tabular}{lllll}
\hline
& $t_s$ [m] & $\mu_{NH}$ [Pa] & $\mu_{fr}$ & $T_{fl}$ [$^{\circ}$C] \\
\hline
\texttt{lower} & 1e-3 & 1e5 & 0.1 & 25 \\
\texttt{upper} & 2e-3 & 1e6 & 1.0 & 35 \\
\texttt{optimal} & 1.57e-3 & 2.80e5 & 0.186 & 29.19 \\
\hline
\end{tabular}
\end{table}

\subsection{Finite element model validation} \label{sec:methods_fe_model_validation}

To validate the calibrated FE model, the net contact force predicted by simulation was compared to corresponding experimental values as illustrated in Figure~\ref{fig:methods_optim_calib}, but now for every possible indenter and trajectory in the datasets. Results of the validation are presented in Section~\ref{sec:results}.

\subsection{Tactile feature estimation} \label{sec:methods_tactile_features}

Two tactile features (i.e., 3D contact location and 3D net force vector) were estimated from BioTac electrode values using neural networks. The \textit{purely experimental} dataset was used. Data from BioTac~1, 2, and/or 3 were used for training and testing; the specific case will be explicitly indicated throughout the Results section. The dataset was split, with approximately 70\% of the indentation trajectories randomly apportioned to training, 20\% to validation, and 10\% to testing. To properly evaluate generalizability, data from each trajectory was kept contiguous during apportionment; in other words, the test set exclusively contained data points from trajectories that were not present in the training set. Input data was normalized to zero mean and unit variance.

Three different network architectures were tested: MLP, 3D voxel-grid-based CNN, and PointNet++. Previous neural-network-based efforts to estimate tactile features utilized MLPs \cite{Wettels2011Robio, Su2015Humanoids} and a 3D voxel-grid-based CNN with sparse occupancy \cite{Sundaralingam2019ICRA}. In addition, our conference publication used PointNet++ \cite{Narang2020RSS}, which directly consumes point cloud data, learns spatial encodings of each point, and hierarchically captures local structure \cite{Qi2017NIPS}. Although conceived for vision, this architecture is conceptually suitable for BioTac electrode values and FE nodal displacements, as these constitute point sets that are sparse, highly nonuniform, or both. Nevertheless, in this paper, all 3 architectures are directly compared to quantitatively determine which is the most suitable for our tactile data.

For the MLP, regressions were performed from electrode values ($\mathbb{R}^{19}$) to contact location ($\mathbb{R}^{3}$) and force vector ($\mathbb{R}^{3}$) independently. No information about electrode locations was provided to the network. The MLP was a fully connected network with layer dimensions [256, 128, 128].

For the 3D CNN, identical regressions were performed. However, information about electrode locations was indirectly provided by discretizing the BioTac into a uniform voxel grid with dimensions [16, 16, 8] and placing the input electrode values at their corresponding locations. The 3D CNN consisted of 2 3D convolutional layers, a 2D convolutional layer, and 2 fully connected layers. The convolutional layers were used for feature extraction, and the fully connected layers were used for regression. Specifically, the 3D convolutional layers had channel dimensions [64, 256], the 2D convolutional layer had 512 channels, and the 2 fully connected layers had channel dimensions [256, 128]. 

For PointNet++, regressions were performed from electrode values and electrode coordinates ($\mathbb{R}^{19 \times 3}$) at each point to contact location and force vector. Thus, information about electrode locations was directly provided. The network consisted of 3 set abstraction layers followed by 2 fully connected layers. The abstraction layers themselves used fully connected layers to extract feature representations from each point and others within a specified radius, and the final fully connected layers were used for regression. Specifically, the first abstraction layer used 3 fully connected layers with dimensions of [64, 64, 128], the second abstraction layer used [128, 128, 256], and the third abstraction layer used [256, 256, 512]. The abstraction layers downsampled the number of points to 32, 8, and 1 by considering points within a radius of $2.5$~$mm$, $7$~$mm$, and $\infty$, respectively. The final 2 fully connected layers had dimensions [256, 128].

All networks were implemented in TensorFlow for our conference publication \cite{Narang2020RSS}. However, for tactile feature estimation in this paper, the networks were reimplemented and refined in PyTorch; thus, minor differences exist between the results presented in the current paper and those presented previously. Training was conducted on an NVIDIA GPU Cloud (NGC) instance.

\subsection{Tactile field estimation} \label{sec:methods_tactile_fields}

FE nodal displacement fields (i.e., displacements of each node of the BioTac skin mesh, relative to their undeformed locations) were also estimated from electrode data using neural networks. Procedures were nearly identical to those used for tactile features. However, the \textit{mixed} dataset (i.e., simulations aligned with experiments) was used. In addition, only the PointNet++ architecture was applied; future work will compare additional network architectures. Specifically, regressions were performed from both electrode values and electrode coordinates to FE nodal displacements ($\mathbb{R}^{N \times 3}$, where $N$ is the number of selected nodes) on the BioTac skin mesh. The skin mesh contained over 4000 total nodes in each simulation; for efficiency, 128 nodes were selected by randomly sampling along the ventral surface of the skin. The original TensorFlow networks from our conference publication were used; thus, the results presented in the current paper are the same as those presented previously.

\subsection{Tactile signal synthesis}

Finally, BioTac electrode values were estimated from FE nodal field data. Again, the \textit{mixed} dataset was used, and the PointNet++ architecture was applied. Specifically, regressions were performed from both FE nodal displacements and FE nodal coordinates (i.e., undeformed locations of each node of the BioTac skin mesh) ($\mathbb{R}^{N \times 3}$) to electrode values. The networks were implemented in PyTorch.

\section{Results and Discussion} \label{sec:results}

\subsection{Finite element model validation}

\begin{figure}[thpb]
  \centering
  \includegraphics[scale=0.99]{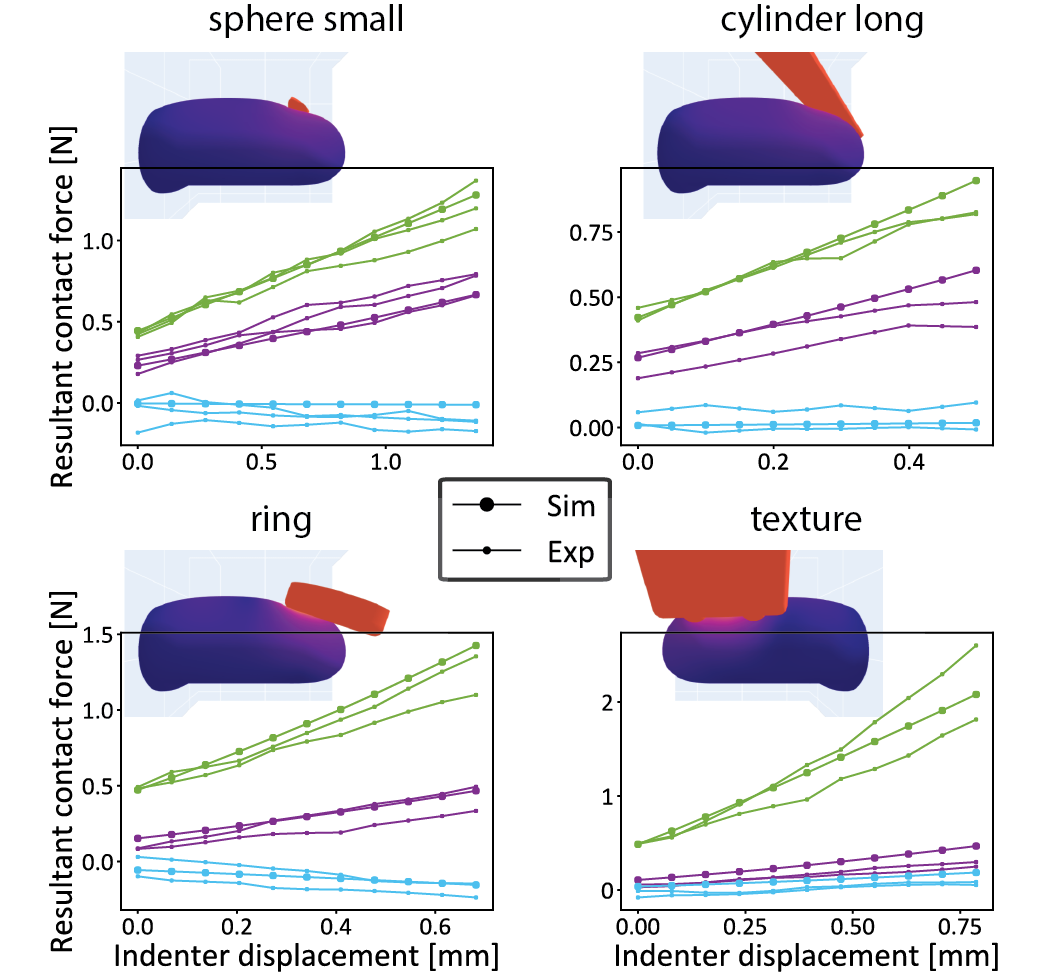}
  \caption{Examples from finite element model validation. Simulation and experimental data are shown for the net contact force acting on the BioTac along each spatial axis. Experimental data is shown for 3 different BioTacs. These comparisons are provided for 4 high-deformation indentation trajectories, each from a randomly-selected indenter. Note that $y$-axis ranges are different across plots; force-deformation derivatives (i.e., stiffnesses) vary strongly with indenter type and contact location.}
  \label{fig:fe_valid}
\end{figure}

As described in Section~\ref{sec:methods_fe_model_calibration}, the free parameters in the FE model were calibrated by minimizing the error between predicted and experimental net forces for a \textit{single} angled trajectory of the \textit{sphere medium} indenter. Following calibration, simulation predictions were compared to experimental data over \textit{all} 9 indenters and 700+ trajectories. For each trajectory, RMS error was computed using Equation~\ref{eq:cost_function}. This process was repeated for all trajectories corresponding to a particular indenter, and the RMS errors were then averaged across the trajectories. The average RMS error ranged from a minimum of $0.226$~$N$ for the \textit{texture} indenter to a maximum of $0.486$~$N$ for the \textit{ring} indenter, with a median of $0.260$~$N$ over all indenters. However, different indenters tend to impart different force levels; when the RMS error for each trajectory was divided by the maximum force magnitude for that trajectory (typically between $1$-$10$~$N$) to compute a \textit{relative} RMS error, the average error was a minimum for the \textit{cylinder small} indenter and a maximum for the \textit{sphere small} indenter. Figure~\ref{fig:fe_valid} compares FE predictions and experimental data for 4 high-deformation trajectories from randomly-selected indenters.

Given the multiple possible sources of experimental variation and error (e.g., BioTac mechanical variability, F/T sensor noise, and registration error), as well as the minimal calibration procedure (i.e., using data from a single indentation out of 700+ possible trajectories), the low RMS error across the wide set of indenters and indentations demonstrates that the FE model is accurate and highly generalizable to novel objects and interactions.

\subsection{Tactile feature estimation} \label{sec:results_tactile_features}

As described in Section~\ref{sec:methods_tactile_features}, 2 tactile features (i.e., 3D contact location and 3D net force vector) were estimated using 3 different network architectures: MLP, 3D voxel-grid-based CNN (referred to in figures as \textit{Voxel}), and PointNet++ (referred to in figures as \textit{PointNet}). For each tactile feature, the following are presented:

\begin{enumerate}
    \item Estimation results for the case of \textit{unseen trajectory} (i.e., the training set and test set both contained data from all indenters, but the test set contained no data points from trajectories that were present in the training set). These results evaluate how well the trained networks can generalize to novel interactions.
    \item Estimation results for the case of \textit{unseen indenter} (i.e., the training set contained data from 8 of the 9 indenters, whereas the test set contained data from only the remaining indenter). Since the BioTac 1 dataset contained fewer indenters (see Section~\ref{sec:methods_exp_testing_procedure}), this case was only examined for BioTac 2 and 3. These results evaluate how well the trained networks can generalize to novel objects.
    \item Estimation results for the case of \textit{unseen BioTac} (i.e., the training set contained data from 1 or 2 BioTacs, whereas the test set contained data from a different BioTac). These results evaluate how well the trained networks can directly transfer to another BioTac.
\end{enumerate}

As a reminder, the networks used to generate results for this subsection were reimplementations of those used in our previous work \cite{Narang2020RSS}; thus, minor differences in numerical values may exist.

\subsubsection{3D contact location}
\begin{figure*}[thpb]
  \centering
  \includegraphics[scale=0.9]{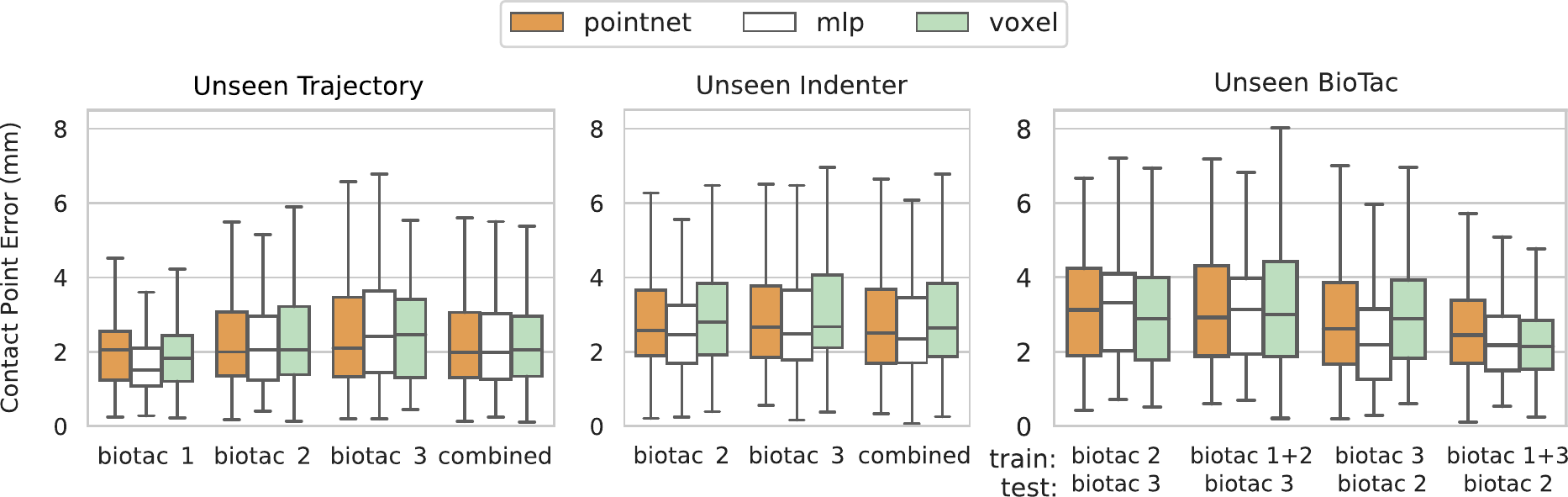}
  \caption{Contact location estimation errors. The test sets consist of unseen trajectories, unseen indenters, or unseen BioTacs. The solid line in each box indicates the median. The box height is equal to the interquartile range (IQR). The fence length is equal to $1.5 * IQR$, rounded down to the nearest data point.}
  \label{fig:results_contact_error}
\end{figure*}

For estimating contact location, all 3 network architectures performed comparably across~\textit{unseen trajectory}, \textit{unseen indenter}, and \textit{unseen BioTac} as shown in Figure~\ref{fig:results_contact_error}. In a small number of instances, MLP performed better than the other architectures. For training and testing with MLP on data from BioTac~1, 2, or 3, as well as data from all BioTacs combined, the median and mean errors in contact location for \textit{unseen trajectory} were \{1.5, 1.8\}~$mm$ for BioTac~1, \{2.0, 2.1\}~$mm$ for BioTac~2, \{2.4, 2.6\}~$mm$ for BioTac~3, and \{2.0, 2.2\}~$mm$ for all BioTacs combined. For \textit{unseen indenter}, median and mean errors for MLP on BioTac~2, 3, and all BioTacs combined were \{2.4, 2.6\}~$mm$, \{2.5, 2.8\}~$mm$, and \{2.4, 2.7\}~$mm$, respectively. For \textit{unseen BioTac}, mean and median errors for MLP are depicted for various cases in Figure~\ref{fig:results_contact_error}.

Comparing to previous works, the best contact location estimates in the literature have errors of $2.4$-$2.9$~$mm$ \cite{Wettels2011Robio}; however, unlike our study, contact location was not controlled or measured with precise instrumentation. Our errors are similar to the spacing between BioTac electrodes ($1.4$-$2.1$~$mm$) and the threshold for human 2-point discrimination at the fingertip ($2.3$~$mm$) \cite{Won2017JApplOralSci}, suggesting that the trained networks may be used for object identification and manipulation of small finger-held objects. Furthermore, as described in Section~\ref{sec:methods_exp_testing_procedure}, the BioTac~2 and 3 datasets contained data for large indenters with distinct edges (e.g., the \textit{cylinder long} and \textit{cube} indenters). Contact location was defined as the \textit{center of the tip} of the indenter, but during angled trajectories, these indenters frequently made initial contact at a corner or edge. Thus, the worst-case median \textit{unseen trajectory} error of $2.4$~$mm$ (i.e., the error for BioTac~3) was substantially better than anticipated, as the networks were occasionally required to infer locations $5$-$10$~$mm$ away.

Comparing the different \textit{unseen} cases, the average median error for all examined datasets was 2.0~mm for \textit{unseen trajectory}, 2.4~mm for \textit{unseen indenter}, and 2.7~mm for \textit{unseen BioTac}. As expected, there was an increase in error from \textit{unseen trajectory} to the more challenging cases of \textit{unseen indenter} and \textit{unseen BioTac}, but the increase was reasonably small. Thus, the trained networks for contact location may generalize to novel objects and BioTacs. At the same time, the interquartile range (IQR) of the error increased by larger proportions (see Figure~\ref{fig:results_contact_error}), implying decreased reliability of the anticipated error magnitude. For \textit{unseen indenter}, the location estimates were nearly as accurate as \textit{unseen trajectory} when the indenter was geometrically similar to others in the training set (e.g., for a \textit{sphere} or \textit{cylinder} indenter), but declined when the indenter was geometrically unique (e.g., for the \textit{ring} indenter, which did not resemble any of the others). Reducing error for \textit{unseen indenter} may require training on an even greater number of indenters; however, reducing error for \textit{unseen BioTac} may be particularly challenging, as the error is likely due to variability in electrical behavior from BioTac to BioTac (see Figure~\ref{fig:data_electrode}).

\begin{figure}[thpb]
  \centering
  \begin{tabular}{c|c|c}
  \includegraphics[trim=200 50 200 50, clip, scale=0.33]{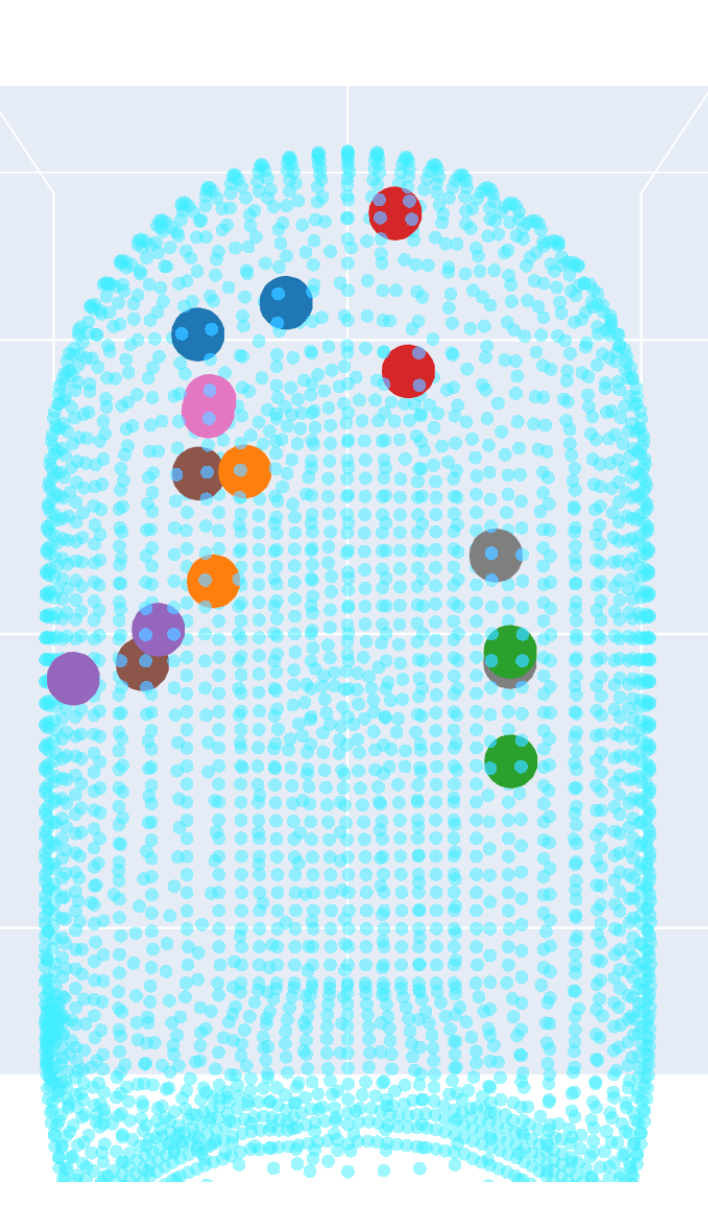} &
  \includegraphics[trim=200 50 200 50, clip, scale=0.33]{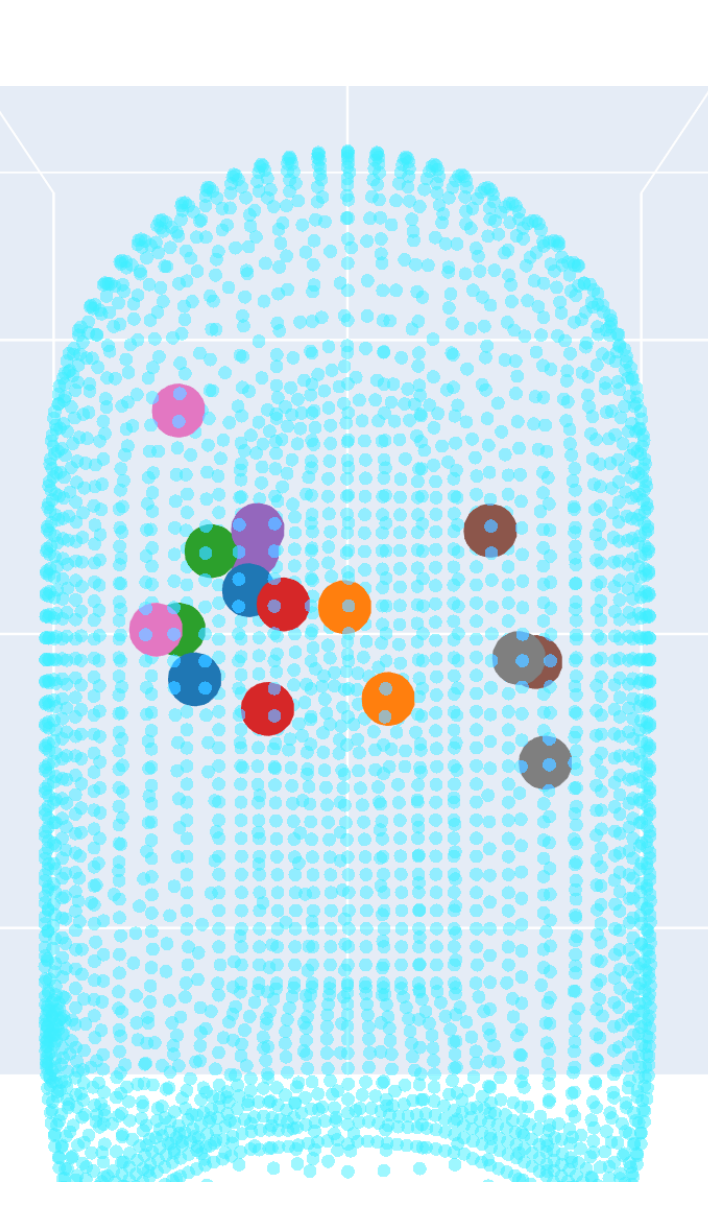} &
  \includegraphics[trim=200 50 200 50, clip, scale=0.33]{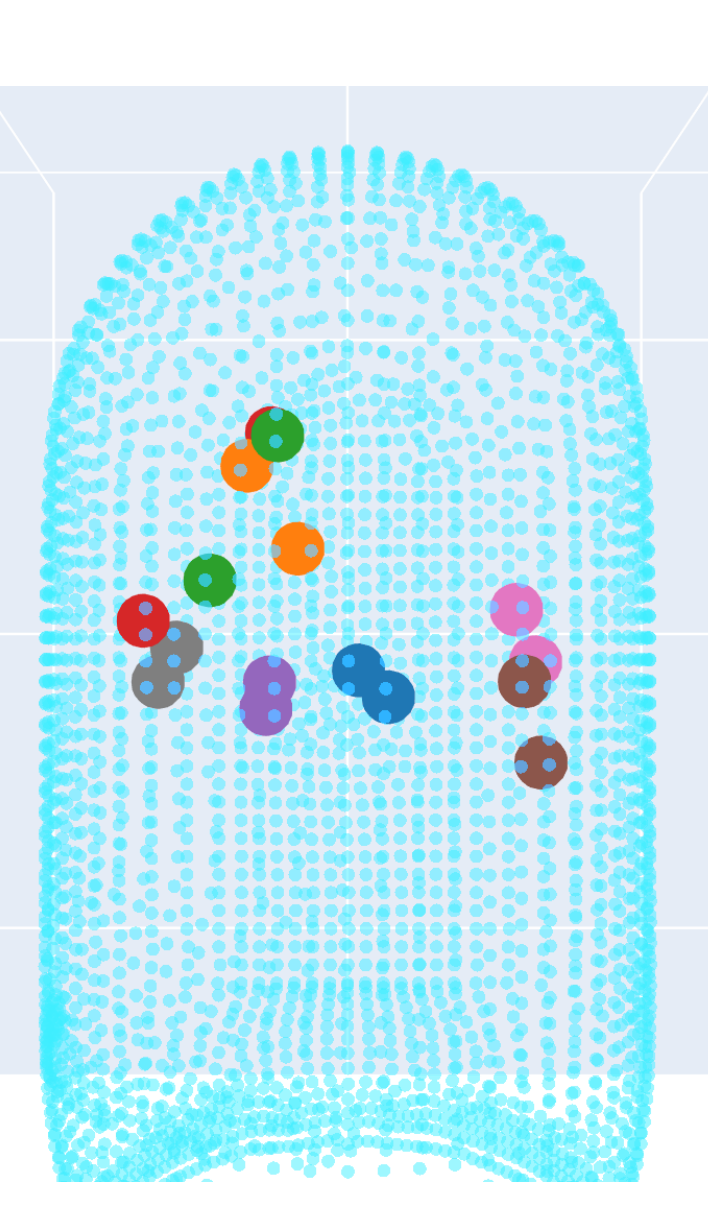}
   \\
  BioTac 2 ($<$2N) & BioTac 2 ($>$8N)     & All Sensors ($>$8N)
  \end{tabular}
  \caption{Examples from contact location estimation. The test sets consist of unseen trajectories. Ventral views of the undeformed BioTac mesh are shown, and same-color points denote target-prediction pairs. Left: 8 randomly-sampled estimates for training and testing on BioTac 2 for low forces. Middle: 8 randomly-sampled estimates for training and testing on BioTac 2 for high forces. Right: 8 randomly-sampled estimates for training and testing on all BioTacs combined for high forces. }
  \label{fig:results_contact}
\end{figure}

Additionally, the hypothesis was examined that increasing indentation intensity can lead to more accurate contact location estimates. The error on the test dataset was computed for 2 different filtering methods: 1) removing all data corresponding to force magnitudes below $8$~$N$, and 2) removing all data corresponding to force magnitudes above $2$~$N$. As shown in Figure~\ref{fig:results_contact}, the high-force dataset has a lower error compared to the low-force dataset. Thus, in real-world applications, contact location estimation may be improved by applying greater force to the object.

\subsubsection{3D net force vector}

\begin{figure*}[thpb]
  \centering
  \includegraphics[scale=0.9]{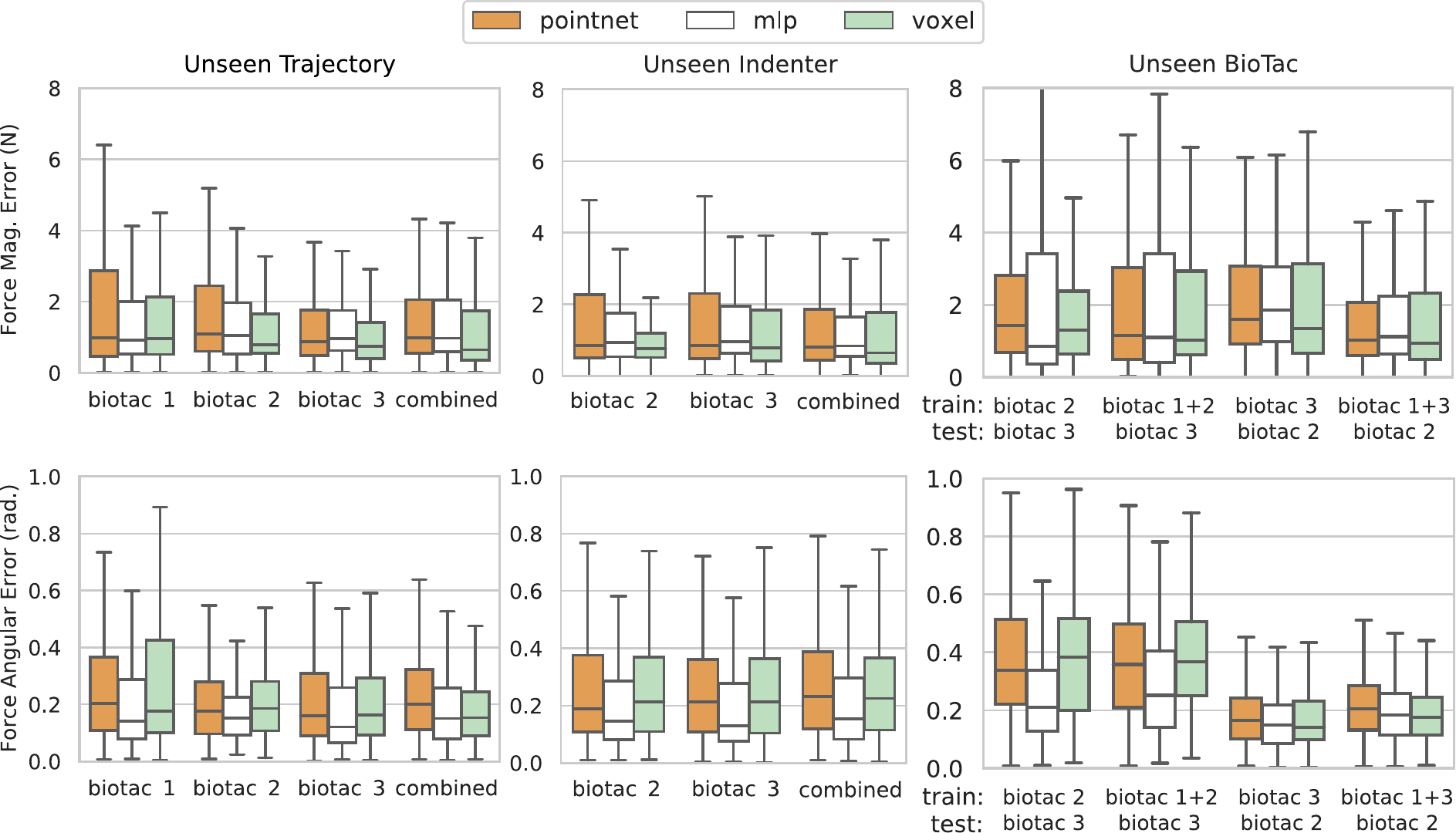}
  \caption{Net force estimation errors. The test sets consist of unseen trajectories, unseen indenters, or unseen BioTacs. Top row: magnitude errors. Bottom row: angular cosine errors. The solid line in each box indicates the median. The box height is equal to the interquartile range (IQR). The fence length is equal to $1.5 * IQR$, rounded down to the nearest data point.}
  \label{fig:results_force_error}
\end{figure*}

\begin{figure}[thpb]
  \centering
  \includegraphics[scale=0.75]{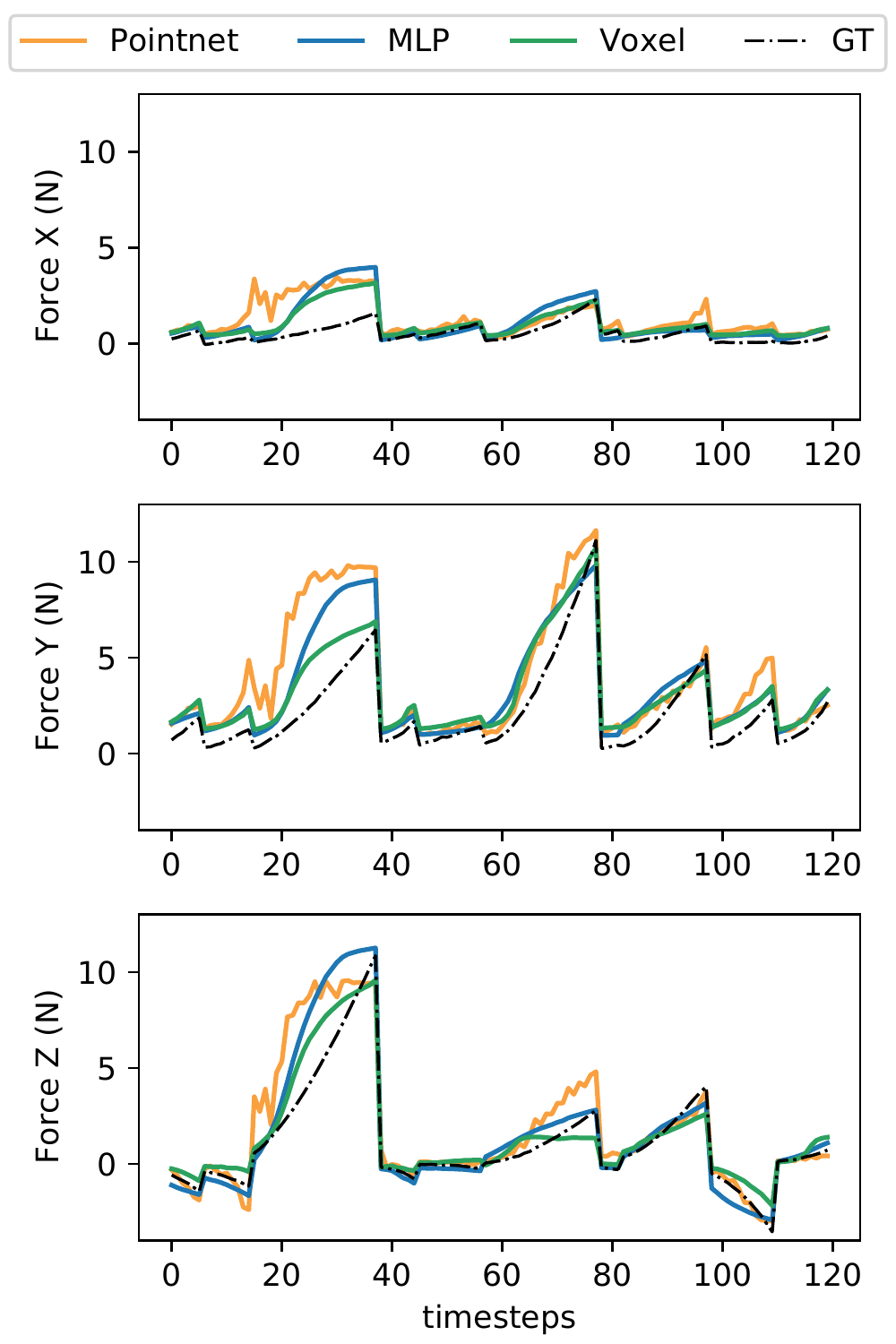}
  \caption{Examples from force vector estimation for unseen trajectories. Force component estimates are shown for a randomly-sampled contiguous slice of multiple trajectories from the test set. The corresponding network was trained and tested on data from a single BioTac. Each spike corresponds to the maximum displacement increment during a particular trajectory. Label \textit{GT} denotes ground truth.}
  \label{fig:results_force}
\end{figure}

For estimating force vector (i.e., magnitude and angle), the best-performing network architecture was 3D voxel-grid-based CNN for force magnitude and MLP for force angle across across \textit{unseen trajectory}, \textit{unseen indenter}, and \textit{unseen BioTac} as shown in Figure~\ref{fig:results_force_error}. For training and testing on the BioTac~1, 2, 3, and combined data, the median errors in force magnitude for \textit{unseen trajectory} were \{0.96, 0.78, 0.73, 0.64\}~$N$, respectively (Figure~\ref{fig:results_force}). Normalized by the standard deviation of the force magnitudes, these values are \{0.22, 0.16, 0.16, 0.14\}. Mean errors tended to be higher due to the high peak force levels examined throughout data collection (i.e., $23$-$32$~$N$). Median angular cosine errors were \{0.14, 0.15, 0.12, 0.15\}~$rad$, respectively.
For \textit{unseen indenter}, median force magnitude errors for \textit{Voxel} on BioTac~1, 2, 3, and all BioTacs were \{0.75, 0.78, 0.64\}~$N$, respectively, and median angular cosine errors for MLP were \{0.14, 0.13, 0.15\}~$rad$. For \textit{unseen BioTac}, mean and median errors for MLP are depicted for various cases in Figure~\ref{fig:results_contact_error}.

Comparing to previous works, the best force estimates in the literature on ground-truth data sources are median force magnitude errors of $0.32$-$0.51$~$N$ and angular cosine errors of $0.07$-$0.36$~$rad$ over smaller peak forces of $1$-$5$~$N$, in the predominantly normal direction \cite{Sundaralingam2019ICRA}. A median force magnitude error of $0.06$~$N$ was also reported, but most training data for that condition was generated from a semi-analytical model. These force estimates were used by \cite{Sundaralingam2019ICRA} to lift and place multiple objects, including a mustard bottle and paper cup. The similarity of our results suggests our estimation framework may also be effective for such applications, with potential robustness to more object types and larger forces, as well as the ability to predict shear.

Comparing network architectures, 3D voxel-grid-based CNN and MLP tended to outperform PointNet++ on the given datasets. \textit{Voxel} and MLP had an order-of-magnitude greater and an order-of-magnitude fewer parameters than PointNet++, respectively, suggesting that performance differences reflected structure rather than capacity. Comparing \textit{Voxel} to PointNet++, the former has a fixed spatial grid, whereas the latter was designed for point clouds with irregular structure; \textit{Voxel} may present advantages for our data, as the electrode locations are fixed in the BioTac coordinate frame. Comparing MLP to PointNet++, the former enables information exchange among all inputs at the shallowest levels of the network; such a structure may be advantageous as well, as the incompressibility of the fluid within the BioTac causes an electrical response at one location of the sensor to immediately induce a complementary response at a distance. Moreover, PointNet++ tended to generate substantially noisier estimates than the other architectures as seen in Fig.~\ref{fig:results_force}. Future work will focus on further investigating these differences.

Comparing the different \textit{unseen} cases, the average median force magnitude and angular cosine errors for all examined datasets were \{0.78, 0.14\} for \textit{unseen trajectory}, \{0.72, 0.14\} for \textit{unseen indenter}, and \{1.1, 0.19\} for \textit{unseen BioTac}. There was no increase in error from \textit{unseen trajectory} to \textit{unseen indenter}, and there was a small increase in error to \textit{unseen BioTac}. Thus, the trained networks for force may generalize across objects and BioTacs. However, as for contact location, the IQR of the error increased by larger proportions (see Figure~\ref{fig:results_force_error}), again implying decreased reliability of the anticipated error magnitude. From a physical perspective, generalization of force may be slightly easier than for contact location, as force is primarily a function of the sum of electrical responses, whereas contact location depends heavily on the shape of the distribution as well.

\subsection{Tactile field estimation} \label{sec:results_tactile_fields}

\subsubsection{Nodal displacement fields}

\begin{figure}[thpb]
  \centering
  \includegraphics[scale=0.99]{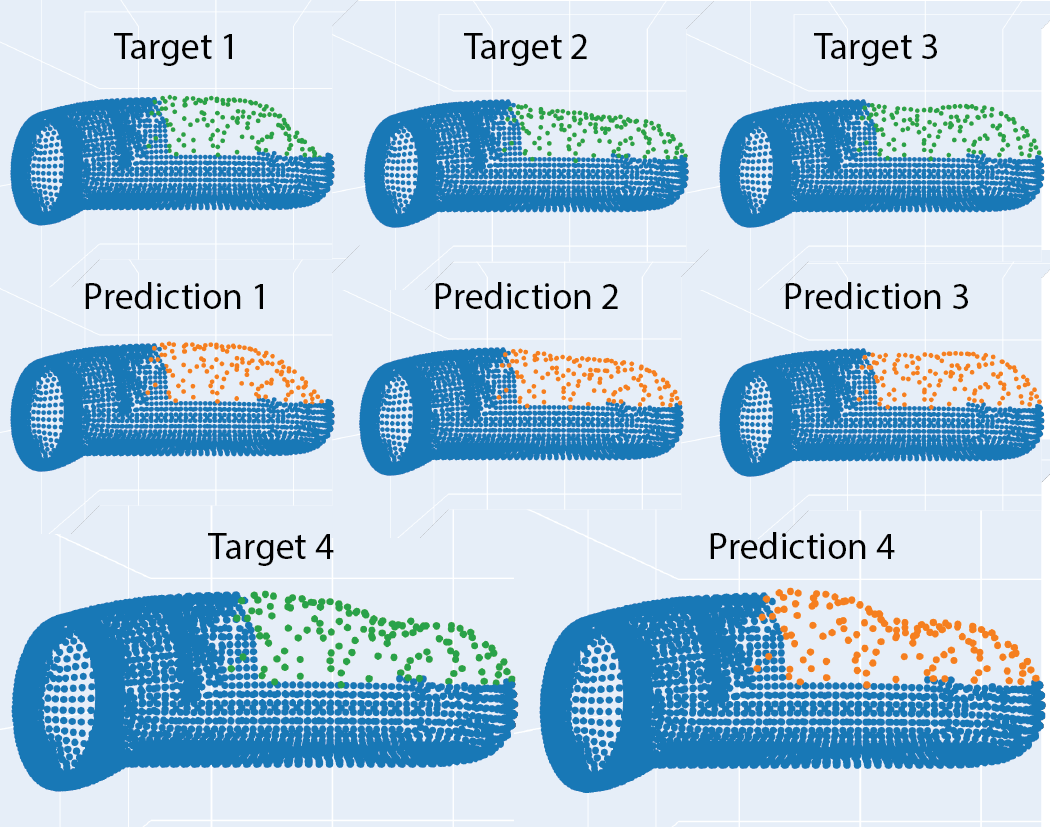}
  \caption{Examples from finite-element nodal displacement estimation. Displacement fields are shown for the maximum displacement increments of randomly-sampled trajectories from the test set. The corresponding network for each target-pair prediction was trained and tested on a distinct subset of the data (i.e., BioTac 1, 2, 3, or combined). Most predicted fields were indistinguishable from targets; several \textit{other} cases are shown here, illustrating minor discrepancies for extreme deformations.}
  \label{fig:results_node_disps}
\end{figure}

As described in Section~\ref{sec:methods_tactile_fields}, nodal displacement fields were estimated using PointNet++. Here, estimation results are presented for the case of \textit{unseen trajectory}; future work will evaluate the cases of \textit{unseen indenter} and \textit{unseen BioTac} as well. For training and testing on the BioTac 1, 2, 3, and combined data, mean nodal displacement errors were \{$0.21$, $0.20$, $0.25$, $0.22$\}~$mm$, respectively. Thus, fingertip deformation fields were well-predicted over the dataset (Figure~\ref{fig:results_node_disps}). Visually, complex deformations were captured, and for moderate-to-high indenter displacements (and thus, strong electrode signals), many deformation fields became indistinguishable from FE predictions.

These results reflect a notable departure from prior studies; rather than a low-dimensional quantity (e.g., center-of-pressure or resultant force vector), high-resolution deformation fields are predicted. From an abstract perspective, the ability to predict these fields supports the long-discussed hypothesis that the BioTac electrode signals do, in fact, contain this high-density information. Practically speaking, such information enables the perception of geometric features such as flat surfaces, edges, corners, and divots, as visible in several of the deformed surfaces of Figure~\ref{fig:results_node_disps}. Furthermore, the generation of these fields provides much greater interpretability of the BioTac's complex electrical response to contact interactions, facilitating the debugging and further development of control algorithms that consume the raw electrode data.

\subsection{Tactile signal synthesis} \label{sec:results_tactile_synthesis}

\subsubsection{Electrode values}

\begin{figure}
    \centering
\includegraphics[scale=0.75]{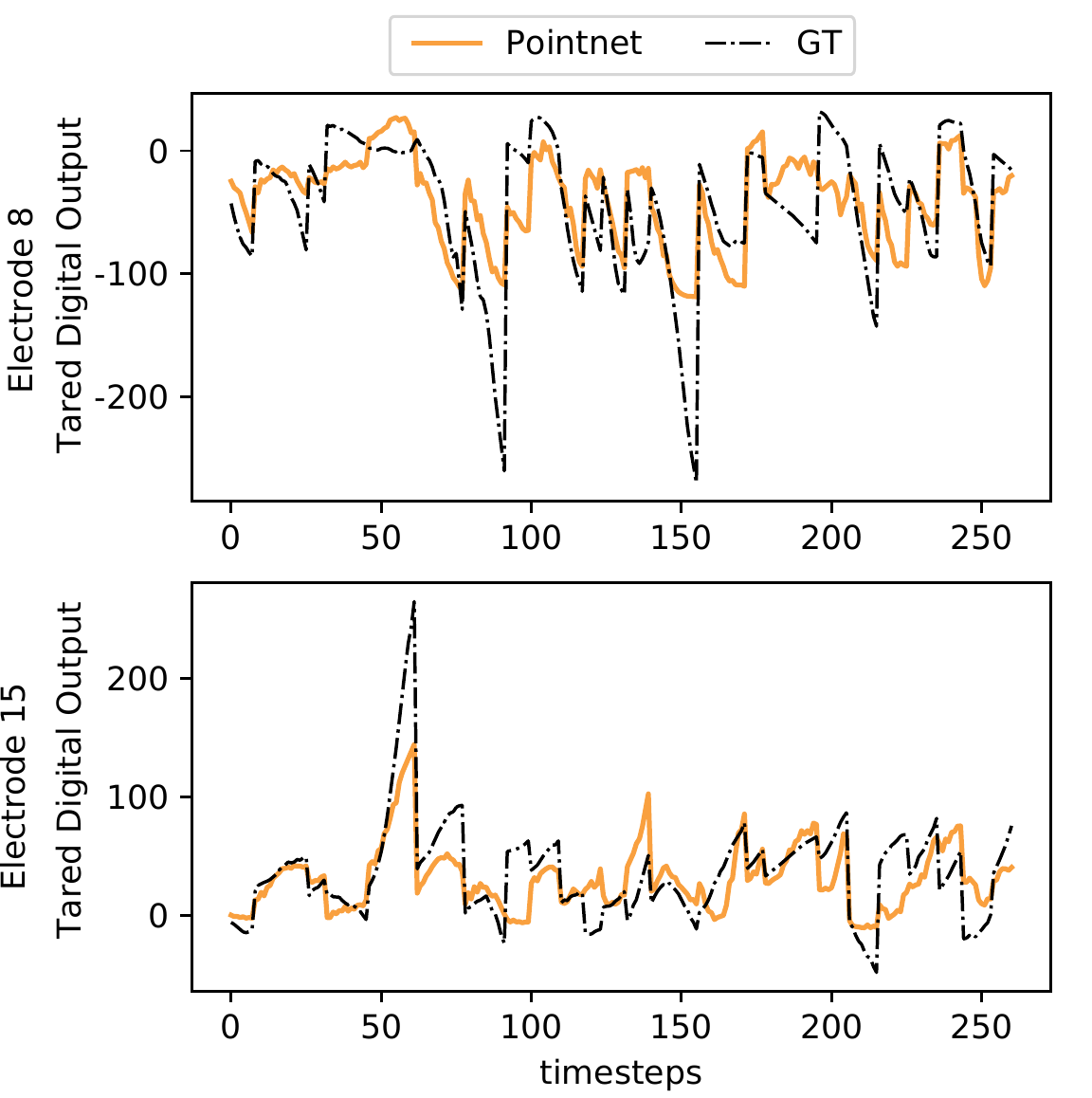} \caption{Examples from electrode value estimation for unseen trajectories. Synthesized electrode values are shown for 2 different electrodes, located at opposite ends of the BioTac (see Figure~\ref{fig:data_electrode}). Label \textit{GT} denotes ground truth. Predicting spikes still presents a challenge, which has been addressed in our concurrent work \cite{Narang2021ICRA}.}
    \label{fig:results_electrode_vals}
\end{figure}

Like the nodal displacement fields, BioTac electrode values were estimated using PointNet++ for the case of \textit{unseen trajectory}. As described in Section~\ref{sec:methods_exp_testing_procedure} and Section~\ref{sec:methods_exp_data_proc}, electrode values were normalized and tared. For training and testing on BioTac 1, 2, and 3, the RMS errors were \{0.019, 0.017, 0.016\} respectively (Figure~\ref{fig:results_electrode_vals}). Converting to raw $12$-$bit$ digital output (0-4096), these values were \{80, 73, 66\}. For the case of \textit{unseen indenter}, the RMS errors for BioTac 2 and 3 were \{0.023, 0.029\}, respectively; the converted digital output was \{98, 120\}.

In comparison, the best electrode value estimates from the literature are an RMSE of 5.9-6.1\% (over the working range of their sensors) for the BioTac SP on seen objects, which corresponded to $195$-$213$ digital output units over their working range \cite{Zapata2020Haptics}, or $242$-$249$ units over our raw $12$-$bit$ range. Note that these estimates were for a dataset that was split non-contiguously; in other words, the training and test sets may have both contained data points from the same trajectories, a case that is less challenging and may limit generalization. (We comment further on contiguity in Section~\ref{sec:results_contiguity}.) On the other hand, even with contiguous samples, our results outperform this benchmark for both seen and unseen objects. For reference, we also report the median $\ell^2$ distance between predicted and ground-truth electrode value arrays ($\mathbb{R}^{19}$) in Table~\ref{tab:results_electrode_error}.
In concurrent work, we have found that using a \textit{less} sophisticated FE simulator improves our results further \cite{Narang2021ICRA}. The new simulator does not impose strict convergence criteria on the simulations, preventing early divergence and allowing FE simulation data to be aligned with experimental data over a greater range of indenter displacements (see procedure in Section~\ref{sec:methods_exp_sim_data_proc}). Thus, the dataset used for synthesizing electrode values can be expanded, enabling more accurate predictions.

\begin{table}[thpb]
\small\sf\centering
\caption{Median $\ell^2$ distance between predicted and ground-truth electrode value arrays ($\mathbb{R}^{19}$) in BioTac digital output range ($0$-$4095$). \label{tab:results_electrode_error}}
    \begin{tabular}{lrrr}
        \hline
        & BioTac~1 & BioTac~2 & BioTac~3 \\
        \hline
        \texttt{Unseen Trajectory} & 225 & 166 & 187 \\
        \texttt{Unseen Indenter} & N/A & 161 & 274 \\
        \hline
    \end{tabular}
\end{table}

\subsection{Data contiguity} \label{sec:results_contiguity}
As described in Section~\ref{sec:methods_tactile_features}, data from each indentation trajectory was kept contiguous when apportioning the training and test datasets (i.e., the test set only contained data from unseen trajectories). As a sanity check, the non-contiguous case was also considered, where individual \textit{data points} were randomly sampled when apportioning the datasets (i.e., the test set contained data points from trajectories that were explored in the training set). As anticipated, estimation accuracy on the test set dramatically improved: mean contact location error decreased by 10x (to approx. $0.2$~$mm$), median force magnitude and angular error decreased by 3x (to approx. $0.15$~$N$ and $0.05$~$rad$, respectively), and mean nodal displacement error decreased by 2.5x (to approx. $0.08$~$mm$). However, in real-world scenarios, one may typically not have access to prior data from trajectories executed at test time. These deceptively-low errors reiterate the need for careful dataset construction when estimating tactile quantities.

\section{Conclusions} \label{sec:conclusions}

\subsection{Summary}

This research advances the interpretation and prediction of robotic tactile signals, focusing on the SynTouch BioTac due to its favorable performance metrics and widespread use in research. The first contribution is a precise, diverse experimental dataset for the BioTac, in which 9 indenters interacted with 3 BioTacs. The dataset consists of over 400 unique indentation trajectories, 800 total trajectories, and 50k data points after subsampling; additionally, approximately 70\% of trajectories were designed to induce shear. The second contribution is the first ever 3D FE model of the BioTac, which accurately predicts the mechanical behavior of the sensor. The model utilizes a hyperelastic material law for the skin, captures fluidic incompressibility, and simulates frictional contact. Despite being calibrated using data from a single indentation, the model generalized strongly across multiple BioTacs, as well as a diverse range of objects and interactions. The model is accompanied by a second dataset that contains the previous experimental data aligned with corresponding FE predictions. The third contribution is a set of neural-network mappings that accurately estimates 3D contact location and force vector, but more importantly, extends the state-of-the-art by predicting tactile \textit{fields} (e.g., nodal displacement fields) from BioTac electrode values. Finally, as outlined in Section~\ref{sec:introduction}, the present extension of our conference publication \cite{Narang2020RSS} adds multiple additional pieces to the work; the most important of these is a set of learned mappings that can synthesize BioTac electrode values.

These results advance robotic tactile sensing for multiple reasons. First, the experimental dataset captures diverse contact interactions beyond existing efforts. For example, previous studies typically focused on manually contacting the BioTac with 1 or 2 indenters, predominantly or exclusively in directions normal to the BioTac surface; the relative pose of the indenter and BioTac was rarely measured with precision, if at all. Nevertheless, most real-world objects have wide geometric diversity, and shear is highly critical for slip detection, inertial estimation, palpation, and sliding or spreading objects. Knowledge of contact location is also invaluable for grasping and dexterous manipulation. Second, as recently initiated for vision-based tactile sensors \cite{Ma2018ICRA, Sferrazza2019Access}, an accurate FE model of the BioTac enables previously-intractable assessments to be performed. Researchers can directly use the model to predict deformation fields of the BioTac and object, as well as contact force distributions transmitted through the BioTac-object interface. These predictions can be leveraged to predict grasp stability with soft contact, anticipate damage to brittle or delicate materials (like eggs, fruit, or living tissue), and guide reshaping strategies for elastoplastic materials (e.g., the flattening of dough). Third, the ability to regress to tactile fields directly from BioTac electrode values enables the high-density information in the FE model to be accessible from raw signals at runtime. Thus, the aforementioned tasks can be accomplished in an online, adaptive, and closed-loop manner. Finally, the ability to synthesize BioTac electrode values enables training of control policies in simulation (e.g., via reinforcement learning) that can accept raw sensor signals as input. Such policies may be transferable to the real world without extensive domain adaptation.

\subsection{Limitations}

The current work has several limitations that present opportunities for future research. The first two limitations are A) the slow speed of the FE simulations, and B) the fully-supervised approach for learning mappings between BioTac electrode values, tactile features, and tactile fields. As described in Section~\ref{sec:methods_fe_model_config}, the FE simulations take an average of $7$~$min$ to execute, making large-scale FE-based assessments (e.g., stability calculations for numerous objects) challenging, and preventing practical use of the FE model within a control loop. In addition, the fully-supervised learning approach demands a carefully-calibrated experimental setup (such as the one in Section~\ref{sec:methods_mech_registration}) to augment the existing dataset with additional data. Fortunately, both of these issues have been addressed in our concurrent work \cite{Narang2021ICRA}, which provides an open-access, GPU-based FE model of the BioTac that is 75x faster, and presents a semi-supervised learning approach that leverages large amounts of unlabeled tactile data to improve the accuracy and generalizability of regressions. 

The second two limitations are A) the decreased accuracy of some network-based mappings when testing on an unseen BioTac, and B) the challenge of transferring our mappings to non-BioTac tactile sensors. As described in Section~\ref{sec:methods_exp_testing_procedure}, Section~\ref{sec:methods_fe_model_validation}, and Section~\ref{sec:methods_tactile_features}, the mechanical behavior of BioTacs is consistent, and the FE model to predict this behavior generalizes strongly across BioTacs. However, the electrical behavior of the BioTac varies strongly from BioTac to BioTac. Although the network-based mappings for contact location and force estimation maintain reasonable accuracy when applied to unseen BioTacs, they may not be accurate enough for highly sensitive applications. In addition, the experimental dataset, FE model, and learned mappings do not transfer naturally to other widely-used tactile sensors, such as the GelSight. For transfer to other BioTacs, a first solution is to complement a network trained on a particular BioTac with additional layers, which can then be fine-tuned for another sensor using a small amount of supervised data (e.g., \cite{Sferrazza2019IROS} for a vision-based sensor). Future work will focus on adapting such techniques to use unlabeled data for the unseen sensor, as well as applying domain randomization during policy training to ensure robustness to transfer error. For transfer to non-BioTac sensors, collecting new experimental data, constructing a new FE model, and retraining networks may be unavoidable requirements. Tactile sensors harness markedly different physical phenomena; furthermore, unlike for cameras, observing an object or the environment with a tactile sensor \textit{changes the behavior of the sensor itself} (e.g., through deformation of a compliant interface). Nevertheless, shared representations of tactile data may exist across sensors of multiple modalities; identification of such representations may greatly facilitate transfer of methods and results from one to another.

Through this work, we present physics-based and data-driven methods to interpret and predict tactile signals for the SynTouch BioTac. By combining physical experiments, FE, and deep learning, we are able to capture both the complex mechanical and electrical behavior of the sensor during real-world interactions. We hope our methods serve as a compelling approach for modeling existing robotic tactile sensors, facilitate training of control policies, and also aid in the design of next-generation devices.

\section*{ACKNOWLEDGMENT}
The authors thank Krishna Mellachervu of ANSYS Inc. for technical support; Keunhong Park of the University of Washington for insightful discussions; Jeremy Fishel of SynTouch Inc. for BioTac CAD models and preliminary mechanical data on the BioTac skin; Miles Macklin of NVIDIA for feedback on finite element modeling; and the RSS reviewers for helpful comments.

\bibliographystyle{./bibliography/IEEEtran}
\bibliography{refs}

\end{document}